\documentclass[11pt]{article}
\usepackage{amssymb}
\usepackage[preprint]{acl}

\usepackage{times}
\usepackage{latexsym}
\usepackage{tabularx}
\usepackage{booktabs}
\usepackage{pifont}
\newcommand{\cmark}{\ding{51}}
\newcommand{\xmark}{\ding{55}}
\usepackage{amsmath}

\usepackage{CJKutf8}
\usepackage{placeins}  
\usepackage{float}
\usepackage{array}

\usepackage{subcaption}
\usepackage{graphicx}
\usepackage[export]{adjustbox} 

\usepackage{tabularx}
\usepackage[T1]{fontenc}

\usepackage[utf8]{inputenc}

\usepackage{microtype}

\usepackage{inconsolata}

\usepackage{graphicx}

\title{Unmasking Reasoning Processes: A Process-aware Benchmark for Evaluating Structural Mathematical Reasoning in LLMs}

\author{
 \textbf{Xiang Zheng\textsuperscript{1,*}},
  \textbf{Weiqi Zhai\textsuperscript{1,*}},
 \textbf{Wei Wang\textsuperscript{1,*}},
 \textbf{Boyu Yang\textsuperscript{1,*}},
\textbf{Wenbo Li\textsuperscript{1}},
\textbf{Ruixiang Luo\textsuperscript{1}},
\\
 \textbf{Haoxiang Sun\textsuperscript{1,2}},
 \textbf{Yucheng Wang\textsuperscript{1}},
 \textbf{Zhengze Li\textsuperscript{1}},
 \textbf{Meng Wang\textsuperscript{1}},
  \textbf{Yuetian Du\textsuperscript{1}},
 \textbf{Guojie Lin\textsuperscript{1}},
\\
 \textbf{Yaxuan Wang\textsuperscript{1}},
 \textbf{Xiaoxiao Xu\textsuperscript{1}},
 \textbf{Yanhu Mo\textsuperscript{1}},
 \textbf{Xuan Ren\textsuperscript{1}},
\textbf{Hu Wei\textsuperscript{1,\textdagger}},
 \textbf{Bing Zhao\textsuperscript{1,\textdagger}}
\\
 \textsuperscript{1}Alibaba Group,
 \textsuperscript{2}Shanghai Jiao Tong University
\\
 \small{
   \textsuperscript{\textdagger}\textbf{Correspondence:}
   \href{mailto:ruiwang@sjtu.edu.cn}{wangrui12@sjtu.edu.cn},
   \href{mailto:xiongdao@alibaba-inc.com}{kongwang@alibaba-inc.com},
   \href{mailto:kongwang@alibaba-inc.com}{xiongdao@alibaba-inc.com}
 }
}

\begin{document}
\maketitle
\footnotetext{
\textsuperscript{*}These authors contributed equally to this work.
}

\begin{abstract}
Recent large language models (LLMs) achieve near-saturation accuracy on many established mathematical reasoning benchmarks, raising concerns about their ability to diagnose genuine reasoning competence. This saturation largely stems from the dominance of template-based computation and shallow arithmetic decomposition in existing datasets, which underrepresent reasoning skills such as multi-constraint coordination, constructive logical synthesis, and spatial inference.
To address this gap, we introduce \textsc{ReasoningMath-Plus}, a benchmark of 150 carefully curated problems explicitly designed to evaluate \emph{structural reasoning}. Each problem emphasizes reasoning under interacting constraints, constructive solution formation, or non-trivial structural insight, and is annotated with a minimal reasoning skeleton to support fine-grained process-level evaluation. 
Alongside the dataset, we introduce \textsc{HCRS} (\textbf{H}azard-aware \textbf{C}hain-based \textbf{R}ule \textbf{S}core), a deterministic step-level scoring function, and train a Process Reward Model (PRM) on the annotated reasoning traces. Empirically, while leading models attain relatively high final-answer accuracy (up to $5.8/10$), HCRS-based holistic evaluation yields substantially lower scores (average $4.36/10$, best $5.14/10$), showing that answer-only metrics can overestimate reasoning robustness.
\end{abstract}

\section{Introduction}


Recent large language models (LLMs) have demonstrated substantial progress in mathematical and logical reasoning~\cite{li2024openai}. Benchmarks such as CMATH~\cite{wei2023cmath}, GAOKAO~\cite{zhang2023evaluating}, ZebraLogic~\cite{lin2025zebralogic}, Hard2Verify~\cite{pandit2025hard2verify}, Omni-MATH~\cite{gao2024omni}, and FineMATH~\cite{liu2024finemath} have played a central role in quantifying these advances and in driving techniques including chain-of-thought prompting, verifier-based reasoning, and reward learning. However, as performance on these benchmarks approaches saturation, it becomes increasingly unclear whether high scores faithfully reflect genuine reasoning competence, particularly the ability to construct structured, minimal, and logically coherent arguments~\cite{wang2024math, qi2024mutual, chen2024alphamath, rafailov2023direct}.


A closer examination reveals that this limitation stems not from a lack of benchmarks, but from systematic biases in how reasoning is evaluated.Existing datasets predominantly emphasize arithmetic computation or competition-style problem solving and focus primarily on final-answer correctness, offering limited insight into the structure and robustness of the reasoning process itself~\cite{paul2024making, prasad2023receval}. Moreover, heavy reliance on public educational resources increases the risk of overlap with pretraining corpora, complicating the distinction between genuine reasoning and memorization~\cite{deng2024investigating, choi2025contaminated, zhang2024careful, carlini2021extracting}. As a result, core reasoning abilities—such as multi-constraint deduction, constructive number-theoretic reasoning, and spatial or diagrammatic inference—remain underrepresented, despite being central to human problem solving~\cite{lin2025zebralogic, beyer2025lexical}.


To address these limitations, we introduce \textsc{ReasoningMath-Plus}, a benchmark of 150 curated problems designed to evaluate \emph{structural} mathematical reasoning with explicit attention to the reasoning process, rather than final-answer correctness alone. The benchmark focuses on cognitively fundamental reasoning skills, including intuitive logic, combinatorial and number-theoretic construction, and spatial reasoning, which have been extensively studied in prior work but are often evaluated in isolation or primarily through answer-level metrics~\cite{cobbe2021training, hendrycks2021measuring, glazer2024frontiermath}. Each problem is annotated with a concise, human-designed \emph{minimal reasoning skeleton} that specifies the essential intermediate assertions required for a correct solution, enabling controlled and comparable process-level evaluation across models. To facilitate cross-lingual analysis and mitigate language-specific biases, we release parallel Chinese and English versions with matched semantics and difficulty.

Beyond final-answer evaluation, we propose a process-level assessment framework tailored to the benchmark. We introduce HCRS (Hazard-aware Chain-based Rule Score), a deterministic scoring function that evaluates reasoning traces via LLM-based step verification, with hazard-based weighting that penalizes earlier errors, consistent with recent step-level verification findings~\cite{pandit2025hard2verify}. We also train a Process Reward Model (PRM) on the annotated reasoning skeletons to provide a learned scoring signal capturing coherence, logical progression, and sufficiency of generated reasoning traces. Together, these scoring mechanisms support fine-grained analysis of reasoning deviations from minimal logical structure and their impact on final correctness.



Taken together, this work introduces a process-aware benchmark for diagnosing structural mathematical reasoning in large language models. Our contributions are summarized as follows:
\begin{itemize}
    \item \textbf{A benchmark for structural reasoning diagnosis.} 
    We curate \textsc{ReasoningMath-Plus}, a collection of 150 problems emphasizing intuitive logic, combinatorial and number-theoretic construction, and spatial reasoning, which are not systematically isolated and diagnosed in existing benchmarks.

    \item \textbf{Human-designed minimal reasoning skeletons.} 
    Each problem is annotated with a concise sequence of essential intermediate assertions (typically 2--10 steps, median 5), providing a task-specific structural reference for controlled step-level analysis without constraining surface realizations.
    
    \item \textbf{A minimal-structure–grounded process evaluation framework.} We introduce HCRS, a hazard-adjusted deterministic scoring function, together with a Process Reward Model (PRM) trained on the same skeleton annotations. By grounding process evaluation in problem-specific minimal reasoning structure, the framework evaluates whether intermediate assertions satisfy essential constraints, rather than only checking the final answer. This reveals substantial gaps between answer-level success and process-consistent reasoning: answer-level scores can reach $5.8/10$, whereas holistic HCRS-based scores average $4.36/10$.

\end{itemize}


\section{Related Work}

\paragraph{Mathematical and logical reasoning benchmarks.} A wide range of benchmarks have been proposed to evaluate mathematical reasoning in large language models. Early datasets such as GSM8K~\cite{cobbe2021training} and ASDiv~\cite{miao2020diverse} focus primarily on arithmetic word problems with short multi-step derivations. These benchmarks have played an important role in demonstrating the effectiveness of chain-of-thought prompting and self-consistency decoding; however, the underlying tasks are highly templated and often overlap with common educational resources, limiting their ability to diagnose deeper reasoning abilities. More advanced benchmarks, including competition-style datasets such as AIME25~\cite{zhang2024american}, HMMT25~\cite{henkel2025mathematician}, and MiniF2F~\cite{zheng2021minif2f}, extend coverage to algebra, geometry, combinatorics, and number theory, capturing higher difficulty levels and longer reasoning chains. Despite this increased difficulty, these benchmarks typically rely on final-answer evaluation and provide full solution traces rather than minimal, structured reasoning representations, which constrains fine-grained analysis of reasoning behavior. 

More recently, efforts such as Omni-MATH~\cite{gao2024omni} and FrontierMath~\cite{glazer2024frontiermath} further expand topical scope and difficulty to university-level and research-oriented problems. While these datasets expose the limitations of current LLMs at the frontier of mathematical reasoning, their heterogeneity in difficulty and reliance on specialized solvers complicate controlled comparison and systematic process-level evaluation. In contrast, \textsc{ReasoningMath-Plus} focuses on a targeted set of structural reasoning skills—multi-constraint logical deduction, constructive combinatorics, and spatial intuition—that remain underrepresented in existing benchmarks. Table~\ref{tab:benchmark_comparison} summarizes key differences between \textsc{ReasoningMath-Plus} and representative datasets.

\begin{table*}[t]
\centering
\small
\setlength{\tabcolsep}{3.5pt}
\renewcommand{\arraystretch}{1.15}

\begin{tabularx}{\textwidth}{
>{\raggedright\arraybackslash}p{3.2cm}
*{5}{>{\centering\arraybackslash}X}
}
\toprule
\textbf{Benchmark}
& \textbf{AIME25}
& \textbf{HMMT25}
& \textbf{C-EVAL}
& \textbf{M3KE}
& \textbf{\textsc{ReasoningMath-Plus}} \\
\midrule
Language
& En
& En
& Zh
& Zh
& Zh \\

Size
& 30
& 30
& 669
& 796
& 150 \\

Problem Type
& Fill-in-the-Blank
& Fill-in-the-Blank / MWP
& MCQ
& MCQ
& Fill-in-the-Blank / MWP \\

Question Len.
& 363.07
& 328.53
& 76.28
& 46.24
& 125.0 \\

Solution Len.
& 2.90
& 10.13
& --
& --
& 213.9 \\

Subject Label
& \cmark 
& \cmark
& \xmark
& \xmark
& \cmark \\

Reasoning Skeleton
& \xmark
& \xmark
& \xmark
& \xmark
& \cmark \\
\bottomrule
\end{tabularx}

\caption{Comparison of \textsc{ReasoningMath-Plus} with representative math reasoning benchmarks.
MWP denotes math word problems and MCQ denotes multiple-choice questions.}
\label{tab:benchmark_comparison}
\end{table*}

\paragraph{Evaluating reasoning processes.} Beyond final-answer accuracy, evaluating the reasoning process itself has become an important research direction~\cite{cobbe2021training}. Early approaches relied on heuristic pattern matching or post-hoc regular expressions to identify invalid reasoning steps~\cite{cobbe2021training, wang2022self}. More recent work employs LLM-based verifiers for step-level evaluation or trains reward models to score reasoning traces, as explored in verifier-guided decoding and PRM-based frameworks~\cite{ouyang2022training, rafailov2023direct}.

In mathematical reasoning, some systems further incorporate symbolic solvers or formal proof assistants to validate intermediate steps~\cite{wei2022chain, glazer2024frontiermath}. While effective for formalized mathematics, such approaches are difficult to generalize to natural-language reasoning tasks, particularly those requiring intuition, construction, or implicit constraints. Consequently, across existing step-evaluation methods, a common limitation is that reasoning steps are typically assessed independently, without accounting for the temporal position of an error within the reasoning chain. Early mistakes often propagate and dominate downstream reasoning, yet this effect is rarely reflected in evaluation metrics. \textsc{ReasoningMath-Plus} addresses this gap by aligning process-level evaluation with human-designed minimal reasoning structure and by explicitly modeling the impact of error position through hazard-adjusted scoring.

\section{Benchmark Curation}

\subsection{Motivation}

\textsc{ReasoningMath-Plus} is designed to expose structural reasoning failures that are often obscured by answer-only evaluation. Rather than increasing symbolic or computational complexity, we focus on problems whose correctness depends on coordinating multiple constraints, eliminating inconsistent possibilities, and constructing minimal yet sufficient reasoning chains. This design enables targeted diagnosis of reasoning errors that are not reliably revealed by existing mathematical benchmarks.

\subsection{Data Collection}

\textbf{Data construction.} \textsc{ReasoningMath-Plus} is curated to elicit \emph{structural} inference rather than symbolic manipulation. Concretely, (S1) we draft candidates via manual authorship and structural adaptation of puzzle-style tasks; (S2) we iteratively refine each item into a precise natural-language statement with explicit constraints and minimal specialized notation; (S3) we check solution well-definedness by independent re-solving and remove or revise candidates with ambiguity or non-unique interpretations; and (S4) we reduce resemblance to common exam/contest templates through targeted rewriting and conservative screening for overly formulaic patterns.

\textbf{Released annotations.} The benchmark consists of 150 problems with explicit process-level annotations. Each item is released with a gold final answer, a full human-written solution, a reasoning skeleton, and subject and difficulty labels. In addition, all problems are provided in parallel Chinese and English versions with aligned semantics and difficulty to support cross-lingual analysis. Table~\ref{tab:reasoningmath_example} presents an example of the annotation schema and reasoning skeleton.

\textbf{Minimal reasoning skeletons.} For process-level evaluation, each problem is annotated with a \emph{minimal} reasoning skeleton---a short sequence of \emph{necessary intermediate assertions} that are sufficient to derive the gold answer. Importantly, \emph{minimal} does not prescribe how a model should write its reasoning: models may produce longer traces with self-verification or deliberation. The skeleton serves as a stable alignment target for step verification by focusing evaluation on essential structural commitments, enabling (i) comparable process scoring across traces of vastly different lengths, and (ii) precise localization of the earliest structural error that drives downstream failure. Skeletons range from 2 to 10 steps (mean 4.65).



\textbf{Subject-level labels for analysis.} In addition to process-level annotations, each problem is assigned a coarse-grained mathematical subject label. These labels are \emph{not} intended to define task formats or target specific domain skills; rather, they serve as an analysis axis that enables comparison with prior mathematical reasoning benchmarks and supports diagnostic breakdowns across familiar categories. We adopt conventional subject labels (algebra, number theory, geometry, combinatorics, and probability) to maintain interpretability and facilitate cross-benchmark analysis, while emphasizing that the primary evaluation focus of \textsc{ReasoningMath-Plus} lies in structural reasoning patterns rather than domain-specific content. In the final dataset, algebra (71 problems) and number theory (51 problems) constitute the majority of instances, reflecting the prevalence of constraint-based deduction and constructive reasoning in these areas. Geometry (12), combinatorics (11), and probability (5) provide additional structural diversity. The subject distribution is summarized in Table~\ref{tab:subject_distribution}.


\begin{table*}[t]
\centering
\small
\setlength{\tabcolsep}{4.5pt}
\renewcommand{\arraystretch}{1.15}

\begin{CJK}{UTF8}{gbsn}
\begin{tabularx}{\textwidth}{l>{\raggedright\arraybackslash}X>{\raggedright\arraybackslash}X}
\toprule
\textbf{Information} & \textbf{Example in Chinese} & \textbf{English Translation} \\
\midrule

\textbf{Question:} & 我设计了一个游戏，给你6个数字，你可以进行加减乘除运算，其中每次加减运算得1分，每次乘除运算得2分，得出指定输出再加六分。
比如我给你 78，2，13，91，1，30，指定输出为 6，请得出一种得分最高的方案。
&
I designed a game where you are given six numbers and may apply addition, subtraction, multiplication, and division.
Each addition or subtraction gives 1 point, each multiplication or division gives 2 points, and producing the target output yields an additional 6 points.
Given the numbers 78, 2, 13, 91, 1, and 30 with target output 6, find a solution with the maximum score.
\\

\textbf{Solution:} &
\textbf{思维链分析标准：} \newline
step1：将题目转化为运算步骤选择问题，每个运算符都有对应分值。 \newline
step2：在保证最终运算结果等于指定输出的前提下，优先使用高分值运算（乘、除）以最大化总分。 \newline
step3：枚举或推导运算符排列组合，计算运算结果与得分，筛选出得分最高方案。 \newline
step4：验证最终方案运算正确且得分满足最大化。 \newline\newline
\textbf{解题分析标准：} \newline
初始化数字集合：78，2，13，91，1，30。 \newline
分析得分规则：加减 = 1 分，乘除 = 2 分，成功得到目标输出额外 +6 分。 \newline
构造表达式：(78 / 2 / 13) + (91 - 1) / 30。 \newline
其中：(78 / 2) = 39（除法，+2 分），(39 / 13) = 3（除法，+2 分）， \newline
(91 - 1) = 90（减法，+1 分），(90 / 30) = 3（除法，+2 分）， \newline
(3 + 3) = 6（加法，+1 分）。 \newline
最终得分：除法 3 次 ×2 分 = 6 分，加减 2 次 ×1 分 = 2 分，额外奖励 6 分，总分 = 14 分。
&
\textbf{Reasoning Skeleton:} \newline
Step 1: Reformulate the task as an operation selection problem where each operator has an associated score. \newline
Step 2: Under the constraint that the final result equals the target value, prioritize high-scoring operations (multiplication and division). \newline
Step 3: Enumerate or derive operator combinations, compute results and scores, and select the highest-scoring valid solution. \newline
Step 4: Verify that the final expression is correct and achieves the maximum score. \newline\newline
\textbf{Detailed Reasoning:} \newline
Initialize the number set: 78, 2, 13, 91, 1, 30. \newline
Scoring rules: addition/subtraction = 1 point; multiplication/division = 2 points; successful target output = +6 points. \newline
Construct the expression: (78 / 2 / 13) + (91 - 1) / 30. \newline
Operations include three divisions and two additions/subtractions, producing the target value 6. \newline
The total score is 14, which is maximal under the given rules.
\\

\textbf{Subject:} & 代数 & Algebra \\
\textbf{Level:} & 难 & Hard \\

\textbf{Answer:} &
使用计算过程 (78 / 2 / 13) + (91 - 1) / 30，最终结果为 6，得分为 14。
&
The expression (78 / 2 / 13) + (91 - 1) / 30 produces the target value 6 with a total score of 14.
\\
\bottomrule
\end{tabularx}
\end{CJK}

\caption{A bilingual example from \textsc{ReasoningMath-Plus}}
\label{tab:reasoningmath_example}
\end{table*}

\section{Methodology}
\label{sec:methodology}
\begin{figure*}[t]
  \centering

  \begin{subfigure}{\textwidth}
    \textbf{(a)}~%
    \includegraphics[width=0.96\linewidth, valign=t, clip, trim=0 0 0 0]{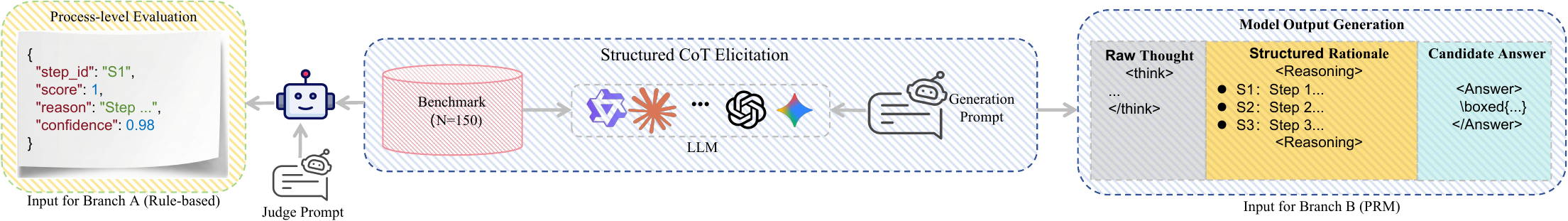}
    \phantomcaption
    \label{fig:sub-a}
  \end{subfigure}

  \vspace{0.2em}

  \begin{subfigure}[t]{0.49\textwidth}
    \vspace{0.3em}
    \textbf{(b)}~%
    \includegraphics[width=0.92\linewidth, valign=t, clip, trim=0 0 0 0]{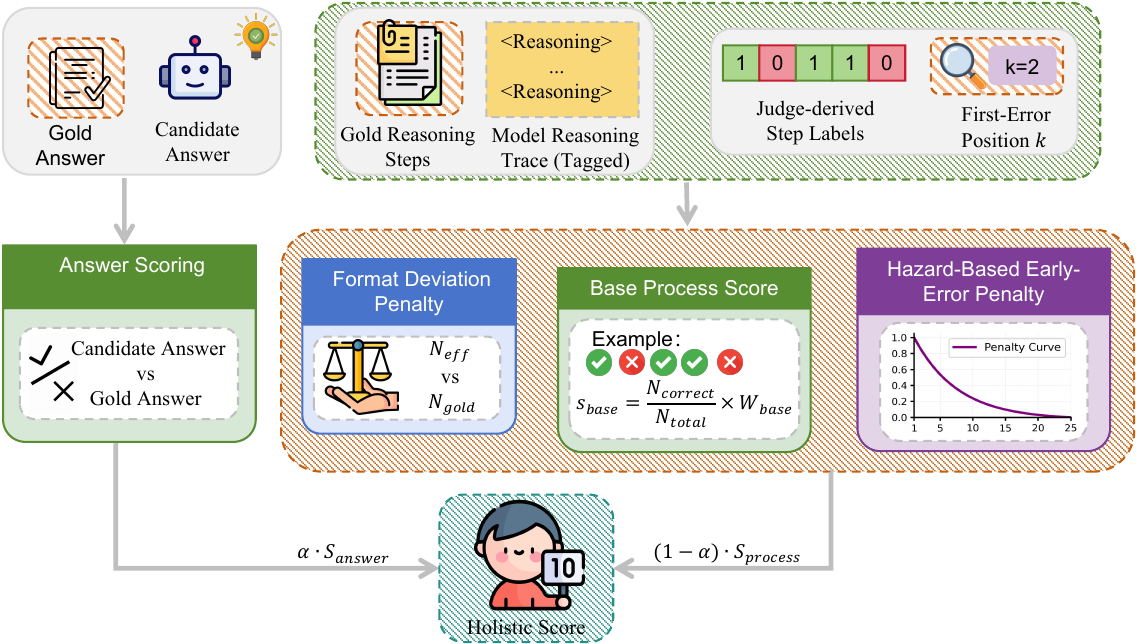}
    \phantomcaption
    \label{fig:sub-b}
  \end{subfigure}%
  \hfill
  \begin{subfigure}[t]{0.49\textwidth}
    \textbf{(c)}~%
    \includegraphics[width=0.92\linewidth, valign=t, clip, trim=0 0 0 0]{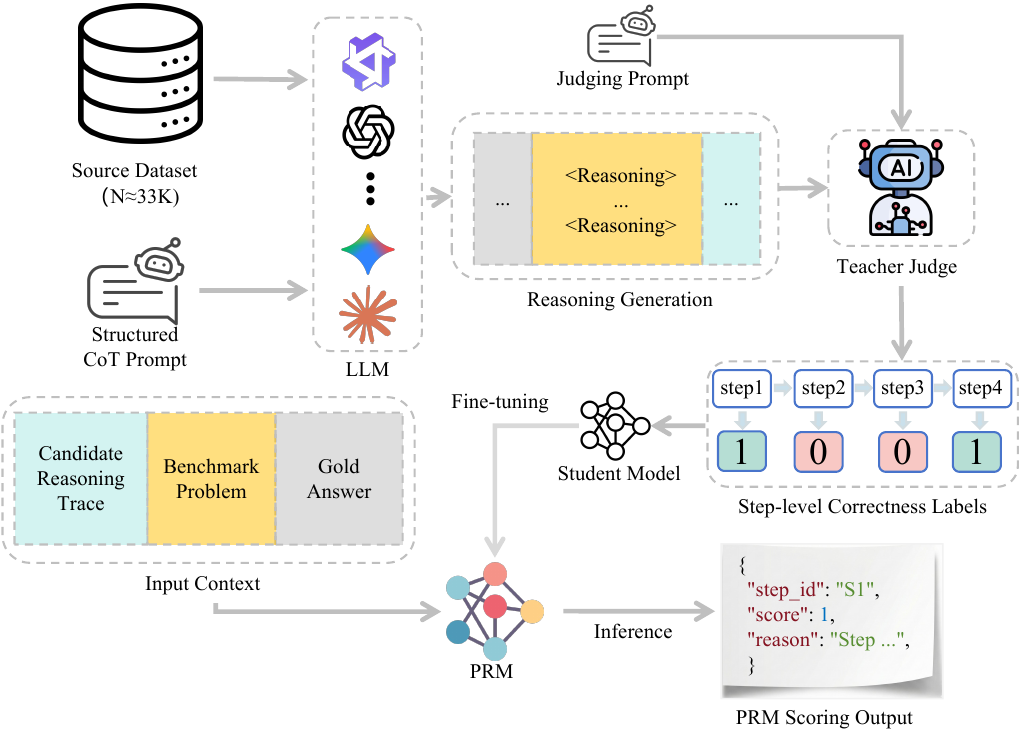}
    \phantomcaption
    \label{fig:sub-c}
  \end{subfigure}
  \vspace{-0.5em}
  \caption{\textbf{Overall framework.}
  \textbf{(a)} Structured CoT elicitation for process-level evaluation.
  \textbf{(b)} Branch A (Skeleton-guided diagnosis): a judge checks step validity against a minimal set of \emph{necessary} skeleton assertions (paraphrase-tolerant; not exact-match), aggregated by HCRS with format and hazard penalties.
  \textbf{(c)} Branch B (Outcome-conditioned verification): a PRM distilled from teacher-judge labels and applied at inference time using only the problem and the gold final answer.}
  \label{fig:overall}
\end{figure*}
To assess reasoning validity beyond outcome correctness, we propose a dual-perspective framework under two supervision regimes (Fig.~\ref{fig:overall}): a \emph{skeleton-guided diagnosis} when dense intermediate annotations are available, and an \emph{outcome-conditioned verifier} otherwise.
\begin{enumerate}
\item \textbf{Branch A: Skeleton-guided Structural Diagnosis (HCRS).}
Given an expert \emph{reasoning skeleton} of necessary assertions, an LLM judge inspects each step for \emph{structural commitment}. This paraphrase-tolerant diagnosis focuses on semantic alignment rather than exact matching. We aggregate results via \textbf{HCRS}, applying format and hazard penalties to strictly penalize early logical fractures.
    \item \textbf{Branch B: Outcome-conditioned Verification (PRM).}
    Without skeleton supervision, a learned Process Reward Model (PRM) verifies each step in an \emph{outcome-conditioned} manner, using only the problem statement and the gold final answer, and outputs discrete step-validity labels for scalable evaluation.
\end{enumerate}
Notably, we show in Section~\ref{sec:prm_eval} that HCRS-style penalties are \textbf{verifier-agnostic} and can improve alignment for both skeleton-guided and outcome-conditioned verifiers.
\subsection{Branch A: Skeleton-guided Diagnosis and HCRS Aggregation}
\label{subsec:stream_a}
\paragraph{Motivation.}
Answer-only evaluation can mask \emph{right answer, wrong reasoning} failures and provides limited granularity for long chains.
We therefore perform \textbf{step-level diagnosis} to localize errors and assign partial credit with explicit justifications.
\paragraph{HCRS Framework.}
Branch A defines \textbf{HCRS}, a deterministic aggregation rule mapping step validity labels from an external judge $J$ to a scalar score via format- and hazard-based penalties.
We instantiate $J$ with \texttt{Gemini-3-Pro} (our fixed teacher judge).
All hyperparameters (e.g., $\alpha,\beta,w$) are fixed across models (Table~\ref{tab:hyperparams}).
\paragraph{Step Validity and Base Score.}
Given an input $x$ and a trace $S_{1:N}$, the judge assigns binary labels $\mathcal{V}=\{v_1,\ldots,v_N\}$ by inspecting each step against skeleton assertions (paraphrases allowed; commitments required).
These labels are then aggregated into a normalized base score:
\begin{equation}
S_{\text{base}} = \frac{10}{N}\sum_{i=1}^{N} v_i.
\end{equation}
\paragraph{Format Deviation Penalty ($P_{\text{fmt}}$).}
Let $r = |N - L_{\text{gold}}| / L_{\text{gold}}$ denote deviation from the reference length $L_{\text{gold}}$.
\begin{equation}
P_{\text{fmt}} = \alpha r e^{\beta r}.
\end{equation}
We apply an asymmetry factor $\eta$ (set $\eta=1.5$ if $N < L_{\text{gold}}$, otherwise $\eta=1.0$) and cap the deduction by $C_{\text{fmt}}$.
\paragraph{Hazard Penalty ($P_{\text{haz}}$).}
For a first error at step $t^\star$, we apply a pre-defined hazard schedule:
\begin{equation}
\tilde{P}_{\text{haz}}(t^\star)=
\begin{cases}
1 - \frac{H(t^\star-1)}{H_{\max}}, & t^\star\le T_{\max}\\
0, & t^\star > T_{\max}
\end{cases}
\end{equation}
and set $P_{\text{haz}}(t^\star)=\min(C_{\text{haz}}, \omega\cdot \tilde{P}_{\text{haz}}(t^\star))$.
The schedule is fixed across all evaluated models (Appendix~\ref{app:prompts}).
\paragraph{Aggregation.}
We define the process-only score as $S_{\text{HCRS}}=\max(0, S_{\text{base}}-P_{\text{fmt}}-P_{\text{haz}})$.
For \emph{reporting only}, we optionally form a holistic score
$S_{\text{hol}} = w S_{\text{HCRS}} + (1-w) S_{\text{ans}}$,
where $S_{\text{ans}}\in\{0,10\}$ and $w=0.7$.
Unless otherwise noted, all analyses use the process-only score $S_{\text{HCRS}}$.
\subsection{Branch B: Outcome-conditioned Verification via PRM}
\label{subsec:stream_b}
Branch B introduces a learned PRM to extend step verification to an \textbf{outcome-conditioned} regime where skeletons are unavailable.
We instantiate the PRM with \textbf{Qwen3-8B-instruct} and train it to predict step validity labels from a fixed teacher judge.
\paragraph{Training Data Construction.}
We build a step-level corpus from \emph{DeepMath}, \emph{OmniThought}, \emph{MiroMind}, \emph{LIMO}, and \emph{NuminaMath}.
From 2{,}500 sampled problems, we generate 35{,}000 candidate traces using 14 LLMs under a unified CoT protocol.
A fixed teacher (\texttt{Gemini-3-Pro}) labels each step with $y_i\in\{0,1\}$ conditioned on the problem and gold final answer.
After filtering malformed outputs, we retain $\sim$33k instances and fine-tune via cross-entropy.
\paragraph{Inference-time Scoring.}
At inference, the PRM acts as a \emph{generative verifier}, producing for each step a rationale $\mathcal{R}_i$ and a discrete validity label $\hat{y}_i\in\{0,1\}$, given by $(\hat{y}_i, \mathcal{R}_i) = f_{\phi}(S_i \mid x, S_{<i})$.
We aggregate labels into a normalized process score:
\[S_{\text{PRM}} = \frac{10}{N}\sum_{i=1}^{N} \hat{y}_i.\]
$S_{\text{PRM}}$ is a process metric and uses the gold answer only through its inclusion in $x$.
\section{Experiments}
\label{sec:experiments}
\subsection{Experimental Setup}
\noindent\textbf{Dataset.}
We evaluate our framework on \textsc{ReasoningMath-Plus}, comprising 150 problems designed to stress long-horizon logical consistency. The dataset emphasizes non-trivial multi-step derivations with unambiguous answers, spanning Algebra, Number Theory, Geometry, Combinatorics, and Probability (see Table~\ref{tab:subject_distribution}).
\noindent\textbf{Models.}
We evaluate 14 endpoints from major families including GPT-5, Gemini, Claude, Grok, Qwen, DeepSeek, and Llama-based models. Precise API identifiers and version strings are detailed in Appendix~\ref{app:model_list}.
\noindent\textbf{Evaluation Protocol and Judge Selection.}
\label{sec:setup_judge}
Models generate structured traces following a fixed schema (\emph{Raw Thought}, \emph{Reasoning Steps}, \emph{Final Answer}).

To ensure rigorous evaluation, we benchmarked multiple candidate judges against human annotations on the generated traces. 
\texttt{Gemini-3-Pro} demonstrated the highest alignment ($R{=}0.64$) and was selected as the global judge for providing step-validity labels in the HCRS pipeline and supervision for PRM training (alignment calibration details in Appendix~\ref{app:judge_selection}).

\begin{figure*}[t]
  \centering
  \begin{subfigure}[t]{0.49\textwidth}
    \textbf{(a)}~%
    \includegraphics[width=0.92\linewidth, valign=t, clip, trim=0 0 0 0]{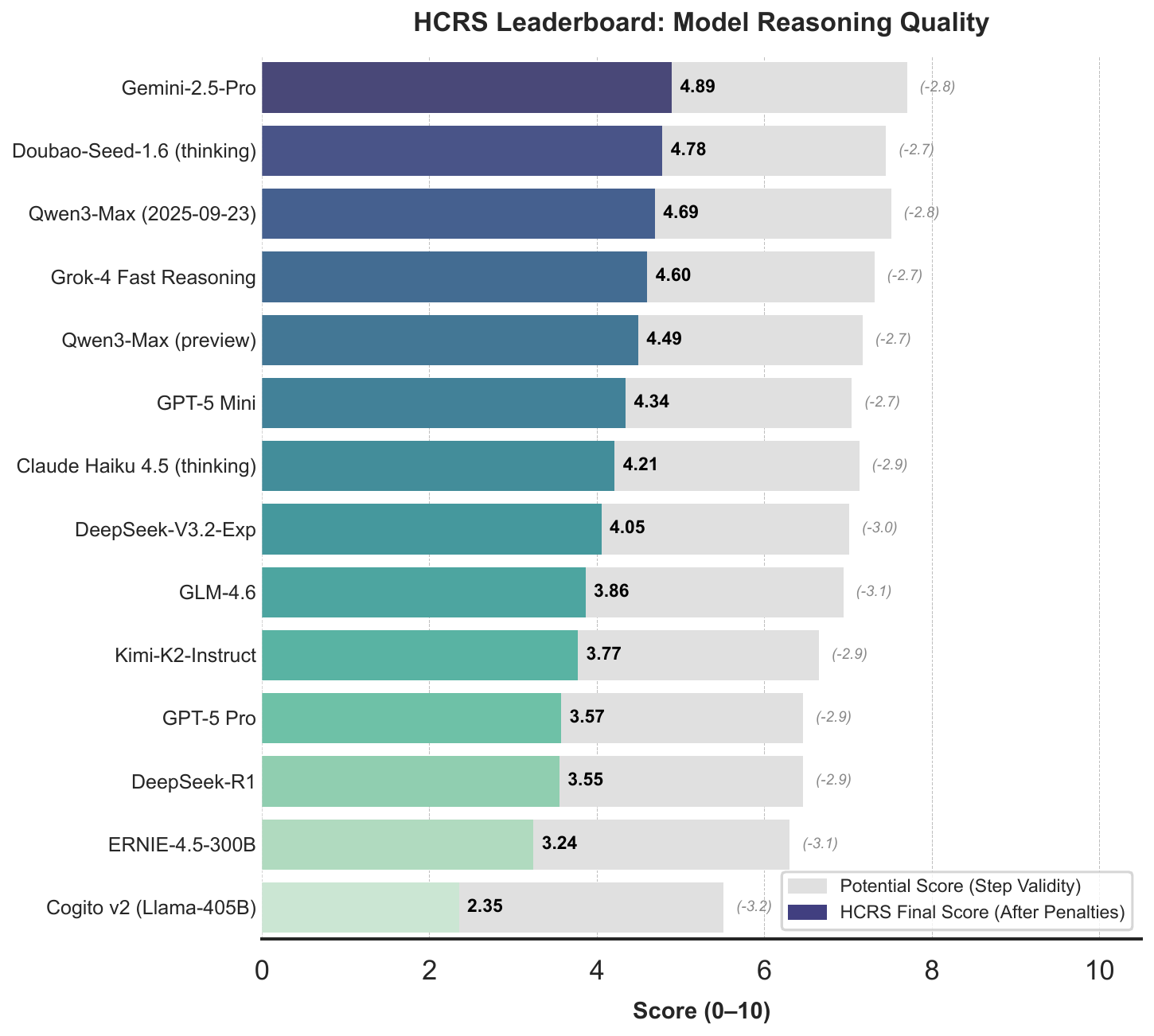}
    \phantomcaption 
    \label{fig:lead-a} 
  \end{subfigure}%
  \hfill 
  \begin{subfigure}[t]{0.49\textwidth}
    \textbf{(b)}~%
    \includegraphics[width=0.92\linewidth, valign=t, clip, trim=0 0 0 0]{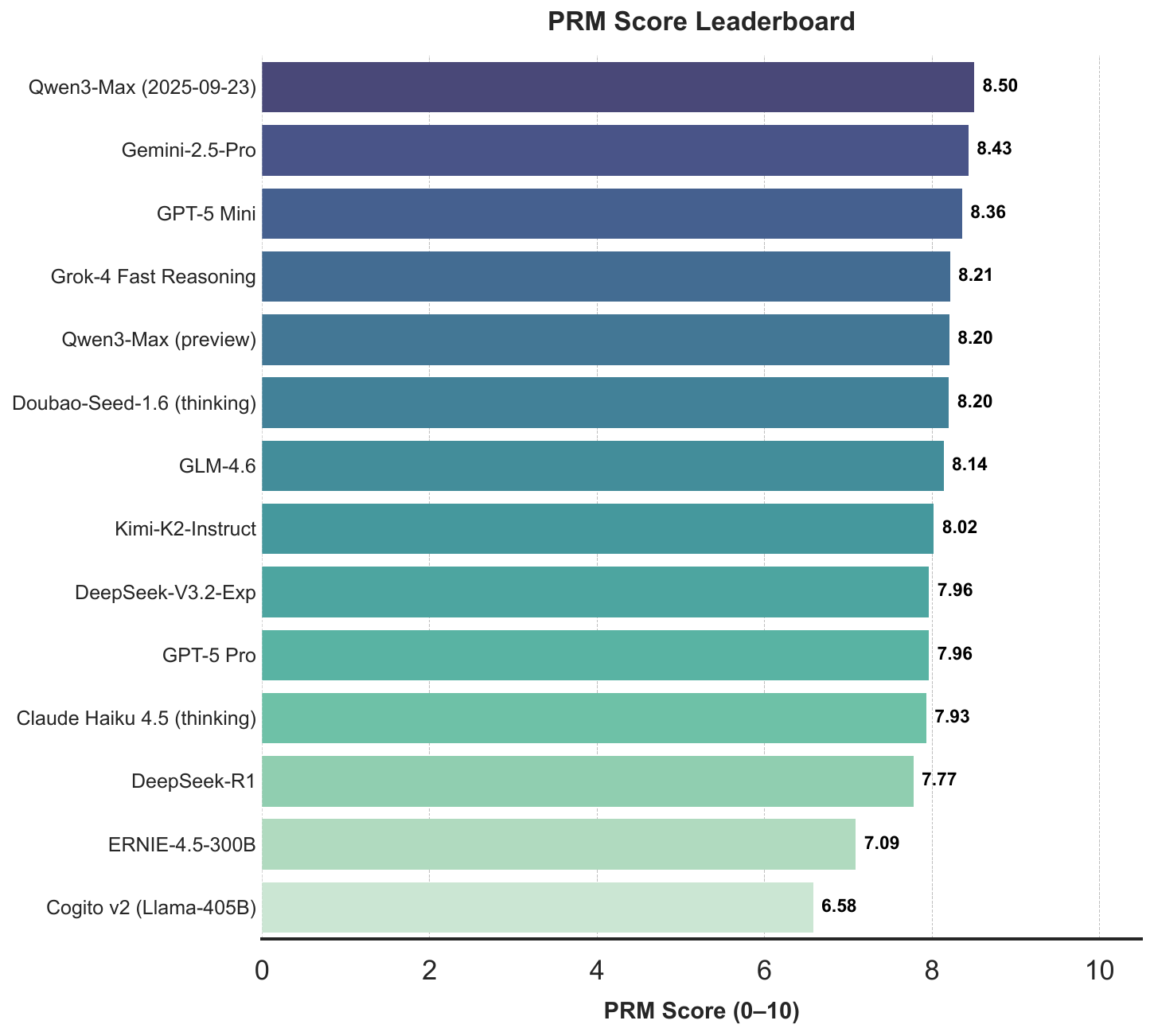}
    \phantomcaption 
    \label{fig:lead-b} 
  \end{subfigure}
  \vspace{1.0em} 
  \begin{subfigure}{\textwidth}
    \centering
    \textbf{(c)}~%
    \includegraphics[width=0.95\linewidth, valign=t, clip, trim=0 0 0 0]{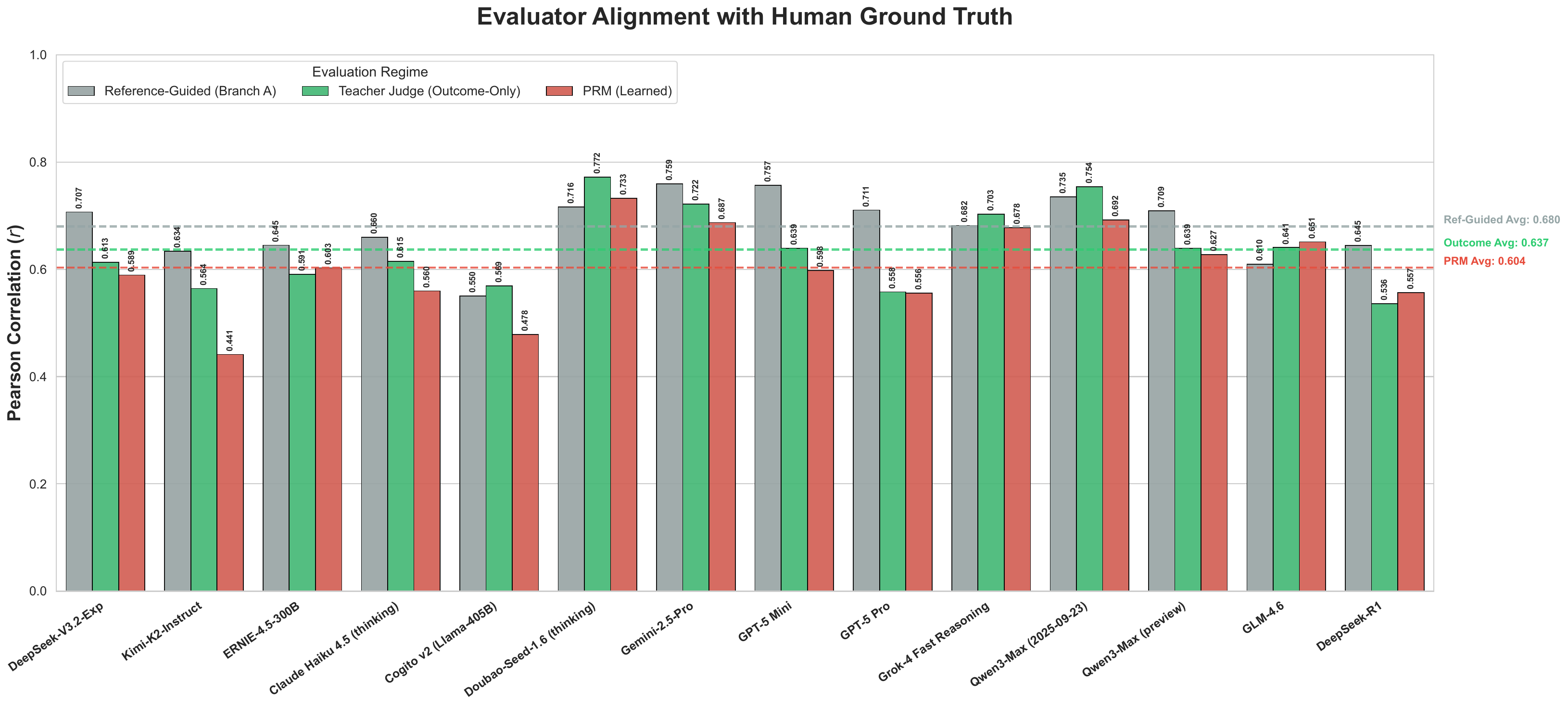}
    \phantomcaption 
    \label{fig:lead-c} 
  \end{subfigure}
  \vspace{-0.5em}
  \caption{\textbf{Evaluation leaderboards and correlation analysis.}
  \textbf{(a)} HCRS leaderboard across domains (grey segments indicate deductions from format and hazard penalties).
  \textbf{(b)} PRM process-score leaderboard under outcome-conditioned step verification.
  \textbf{(c)} Pearson correlation of evaluation methods (Reference-guided, teacher judge, and PRM) against human judgments.}
  \label{fig:leaderboard_comparison}
\end{figure*}

\begin{figure*}[!t]
  \centering
  \includegraphics[width=0.95\textwidth]{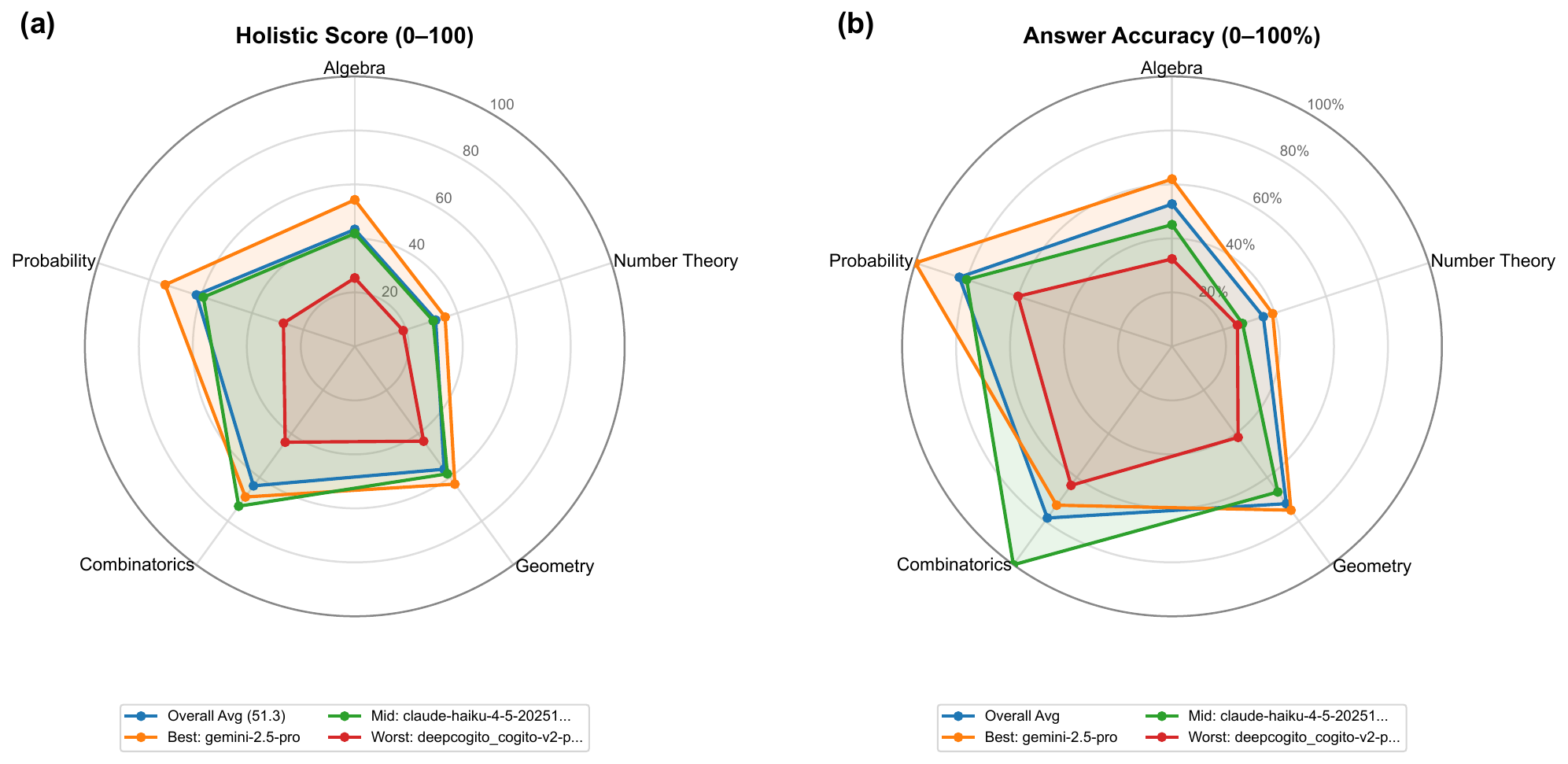}
  \caption{\textbf{Subject-wise capability analysis.}
  Radar plots comparing \textbf{(a)} Holistic Score (0--100) and \textbf{(b)} Answer Accuracy (0--100\%) across five domains.
  Curves correspond to the \textbf{Best}, \textbf{Median}, and \textbf{Worst} models selected by overall Holistic Score, together with the overall average.}
  \label{fig:radar_analysis}
\end{figure*}

\subsection{Main Results}
\label{sec:main_results}

We present our experimental findings organized by the two evaluation protocols defined in Section~\ref{sec:methodology}.
First, we analyze model robustness and structural fragility using the rule-based \textbf{HCRS} (Sec.~\ref{sec:auto_eval}).
Second, we examine semantic reasoning quality using the learned \textbf{PRM} (Sec.~\ref{sec:prm_eval}).

\subsubsection{Structural Diagnosis via HCRS}
\label{sec:auto_eval}


\paragraph{HCRS Leaderboard and Penalty Gaps.}
Figure~\ref{fig:lead-a} shows that HCRS induces a clearer stratification than answer accuracy alone.
Gemini-2.5-Pro ranks first (4.89), followed by Doubao-Seed-1.6 (thinking) (4.78) and Qwen3-Max (2025-09-23) (4.69).
The grey bars denote deductions from format deviations and the hazard-based \emph{first-error penalty}.
These penalties correspond to a substantial reduction of approximately $2.7$--$3.2$ points, amounting to roughly 30\% of the 10-point scale.
This divergence between potential and realized reasoning quality also appears at the instance level (Figure~\ref{fig:lucky_guess_dist}).

\paragraph{Quantifying ``Lucky Guesses''.}
Figure~\ref{fig:lucky_guess_dist} shows the process score distribution for the 996 samples where the final answer is correct. 
We observe that \textbf{6.63\%} (66/996) of these instances fall into the low-score range ($S_{\text{HCRS}}{\le}3$). 
This indicates that while traditional outcome-based metrics would classify these samples as correct, our structural evaluation identifies them as reasoning failures (i.e., ``lucky guesses'').

\paragraph{Domain-wise diagnostic breakdown (descriptive).}

Figure~\ref{fig:radar_analysis} provides a diagnostic comparison between composite scores (a) and answer accuracy (b).
While limited by small sample sizes in sub-domains like Probability and Combinatorics (see Table~\ref{tab:subject_distribution}), the radar charts reveal a critical divergence.
We observe a marked \emph{inward contraction} for mid-tier models (e.g., the green trajectory) when moving from accuracy to composite scores, particularly in combinatorial tasks.
This indicates that a portion of their apparent accuracy masks brittle or formatted-invalid reasoning.
Conversely, \texttt{Gemini-2.5-Pro} exhibits the most stable envelope across both figures, demonstrating that its high performance is supported by rigorous step-wise validity rather than lucky guesses.

\subsubsection{Alignment with Human Judgments}
\label{sec:human_alignment}

We evaluate the reliability of our scoring methods by measuring their consistency with human-annotated scores.
First, we examine linear alignment using Pearson correlation ($R$), as shown in Figure~\ref{fig:lead-c}.
Despite operating in an \emph{outcome-conditioned} setting (i.e., conditioned only on the problem and the gold final answer, without access to gold reasoning steps), PRM achieves a competitive average correlation of $R=0.602$.
This closely tracks the Pro-tier teacher judge (\texttt{Gemini-3-Pro}, $R=0.639$).

To assess robustness beyond linear correlation, we further report rank-aware and agreement-based metrics in Appendix Figure~\ref{fig:agreement_analysis}, including Spearman's $\rho$ (monotonic rank correlation), Kendall's $\tau$ (pairwise concordance), and Quadratic Weighted Cohen's $\kappa$ (agreement intensity).
These complementary metrics provide a more comprehensive view of evaluator reliability beyond Pearson correlation alone.

Crucially, PRM (Method C) remains competitive across all metrics ($\rho=0.684$, $\tau=0.565$, $\kappa=0.568$), confirming that it largely preserves the relative quality rankings preferred by humans even without access to gold intermediate steps.

\subsubsection{Scalable Verification via PRM}
\label{sec:prm_eval}

Figure~\ref{fig:lead-b} presents the PRM rankings. While the top-tier hierarchy mirrors HCRS (with Qwen3-Max and Gemini-2.5-Pro leading), the score distribution exhibits a distinct \textbf{compression effect}.
Unlike the steep $\sim$52\% performance drop observed in HCRS, the lowest-ranked model in the PRM leaderboard retains $\sim$77\% of the top score (6.58 vs. 8.50).
This saturation stems from the PRM's averaging-based aggregation ($\frac{1}{N}\sum \hat{y}_i$), which grants partial credit for locally valid steps even within flawed chains. Consequently, the PRM functions as a smoother measure of \textbf{local semantic consistency}, complementing the strict structural stratification of HCRS.
\subsubsection{When Simple Rules Improve Verifiers}
A key observation is that the HCRS penalty terms can serve as a simple, verifier-agnostic refinement on top of raw step-wise scoring signals.
As shown in Figure~\ref{fig:ablation_study}, applying the same HCRS penalties (\emph{format penalty} and \emph{first-error penalty}) to both the teacher judge (\texttt{Gemini-3-Pro}) and the trained PRM consistently improves alignment with human judgments.
Notably, the gain is more pronounced for PRM, whose average Pearson correlation increases from $0.604$ to $0.633$ (vs.\ $0.637$ to $0.642$ for Gemini).
This suggests that rule-based diagnosis captures systematic structural failure modes that are not fully reflected by raw step-wise validity predictions, providing a generalizable enhancement for process evaluation even in settings where expert skeletons are unavailable.
\section{Conclusion}
\label{sec:conclusion}
In this work, we addressed the opacity of outcome-based evaluation by introducing a dual-perspective framework for long-horizon mathematical reasoning.
Our approach integrates \textbf{HCRS}—a \emph{skeleton-guided} protocol for explicit structural diagnosis—with a \textbf{PRM} designed for \emph{outcome-conditioned} semantic verification.
Experiments on our curated benchmark reveal that answer-only metrics significantly overestimate reliability by masking "lucky guesses," a phenomenon effectively quantified by our hazard-aware penalties.
Validated by high alignment with human experts, this framework bridges the gap between high-precision structural \textbf{diagnostic} signals and flexible learned verification, establishing a transparent and scalable paradigm for verifiable reasoning.
\newpage
\section{Limitations}
\label{sec:limitations}
\paragraph{Limitations.}
Our framework targets fine-grained, process-level auditing and therefore involves modest practical overhead.
In particular, constructing reasoning skeletons and gold step counts may require expert effort, which can increase annotation cost compared to outcome-only evaluation.
That said, this design choice enables more precise diagnosis of step-wise consistency and error propagation, and can serve as a high-quality supervision source for training lighter-weight verifiers in future work.

\hspace{1cm}
\bibliography{custom}

\appendix
\section{Data Statistics}
\label{app:data_stats}
\begin{table}[h]
\centering
\small
\begin{tabularx}{\linewidth}{l X r}
\toprule
\textbf{Subject Category} & \textbf{Category Content} & \textbf{Size} \\
\midrule
Algebra 
& Equation and inequality reasoning, functional relationships, algebraic constraints, and symbolic abstraction. 
& 71 \\

Number Theory 
& Divisibility, parity arguments, modular arithmetic, and constructive reasoning over integers. 
& 51 \\

Geometry 
& Planar and spatial configuration reasoning, geometric relationships, and transformations. 
& 12 \\

Combinatorics 
& Counting arguments, permutations and combinations, case analysis, and discrete construction. 
& 11 \\

Probability 
& Reasoning about random events, conditional probability, and basic probabilistic inference. 
& 5 \\
\bottomrule
\end{tabularx}
\caption{Subject categories and distributions of \textsc{ReasoningMath-Plus}.}
\label{tab:subject_distribution}
\end{table}
\section{Judge Model Selection and Calibration}
\label{app:judge_selection}

\begin{figure}[h]
    \centering
    \includegraphics[width=1.0\linewidth]{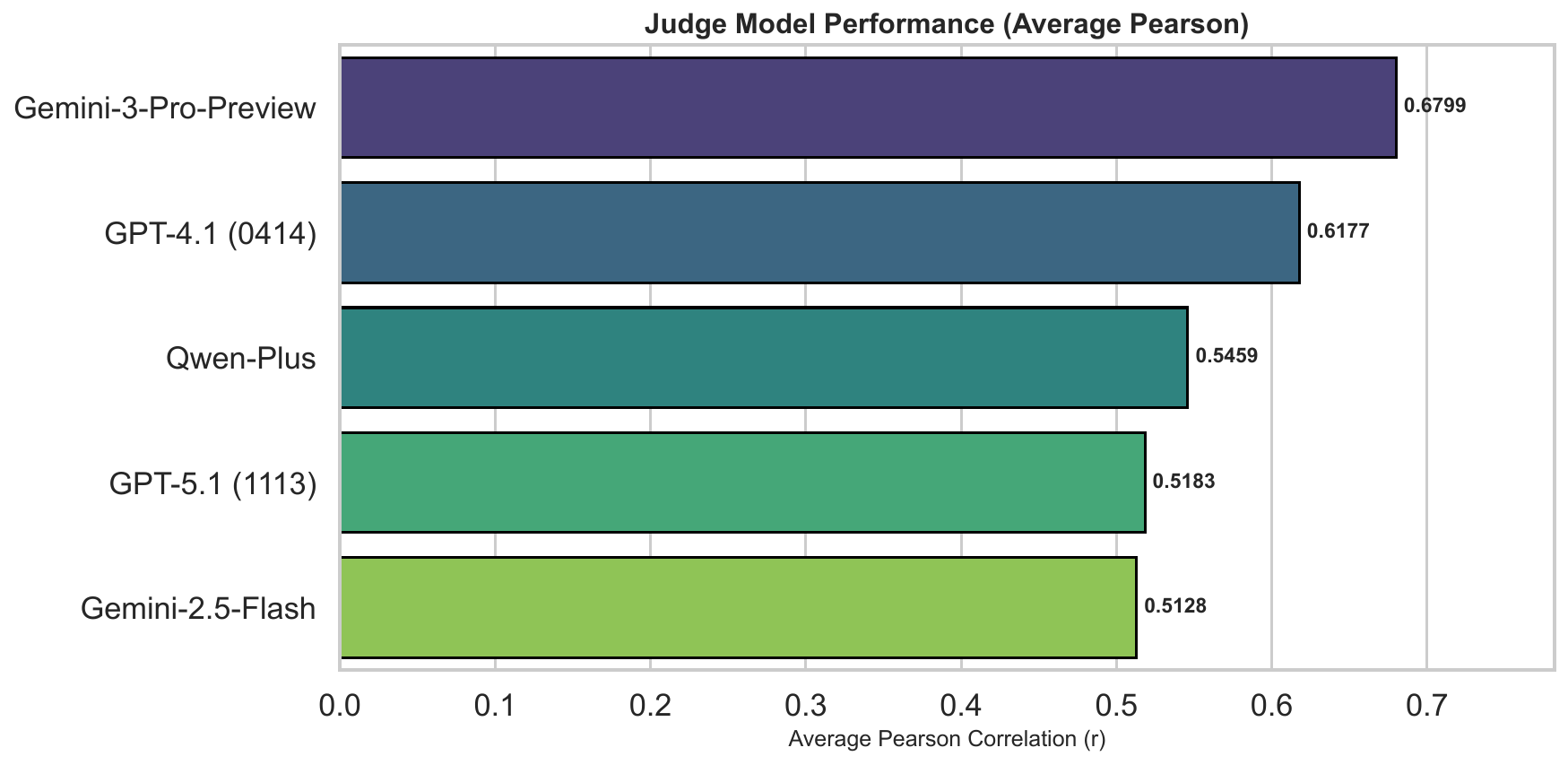}
    \caption{\textbf{Judge performance comparison.} Comparison of candidate judge models based on average Pearson correlation with human annotations ($N{=}2,100$). \texttt{Gemini-3-Pro} exhibits the strongest alignment ($R{=}0.64$).}
    \label{fig:judge_performance}
\end{figure}

To ensure the reliability of our automated evaluation instrument, we performed a rigorous calibration study prior to the main experiments. This section provides the technical details of the judge selection process that were omitted from the main text for brevity.
\paragraph{Data Sampling and Diversity.}
The calibration dataset comprises a comprehensive set of $2{,}100$ reasoning traces ($150 \text{ problems} \times 14 \text{ models}$) generated during the benchmark's preliminary phase. This large-scale sampling allows candidate judges to be evaluated across a diverse spectrum of reasoning behaviors, ranging from concise logical derivations to verbose chain-of-thought explorations. Furthermore, the dataset spans a wide array of trace qualities---from perfect derivations to complete logical collapses---while accounting for varying levels of format adherence to ensure the judge's robustness against minor structural deviations.
\paragraph{Candidate Judges and Methodology.}
We benchmarked several state-of-the-art LLMs as candidate judges, including \texttt{GPT-4o}, \texttt{Claude-3.5-Sonnet}, and \texttt{Gemini-3-Pro}. For each candidate, we applied the reference-guided prompt (Branch A) to generate step-level validity labels. These labels were then aggregated via the HCRS pipeline to produce process-level scores.
\paragraph{Alignment Metrics and Selection Result.}
The primary metric for selection was the \textbf{Pearson correlation ($R$)} between the judge-generated HCRS scores and ground-truth scores provided by human experts. As shown in Figure~\ref{fig:judge_performance}, \texttt{Gemini-3-Pro} demonstrated the highest consistency with human judgments, achieving a correlation coefficient of $R{=}0.64$. Crucially, as noted in Section~\ref{sec:setup_judge}, this calibration involves no tuning of the evaluated models or scoring rules; the chosen judge is fixed globally for all subsequent analyses to maintain the integrity of the evaluation.
\paragraph{Candidate Judges and Methodology.}
We benchmarked several state-of-the-art LLMs as candidate judges, including \texttt{GPT-4o}, \texttt{Claude-3.5-Sonnet}, and \texttt{Gemini-3-Pro}. For each candidate, we applied the reference-guided prompt (Branch A) to generate step-level validity labels. These labels were then aggregated via the HCRS pipeline to produce process-level scores.
\paragraph{Alignment Metrics and Selection Result.}
The primary metric for selection was the \textbf{Pearson correlation ($R$)} between the judge-generated HCRS scores and ground-truth scores provided by human experts. As shown in Figure~\ref{fig:judge_performance}, \texttt{Gemini-3-Pro} demonstrated the highest consistency with human judgments, achieving a correlation coefficient of $R{=}0.64$. 
Crucially, as noted in Section~\ref{sec:setup_judge}, this calibration involves no tuning of the evaluated models or scoring rules; the chosen judge is fixed globally for all subsequent analyses to maintain the integrity of the evaluation.

\begin{figure*}[h]
    \centering
    \includegraphics[width=0.95\linewidth]{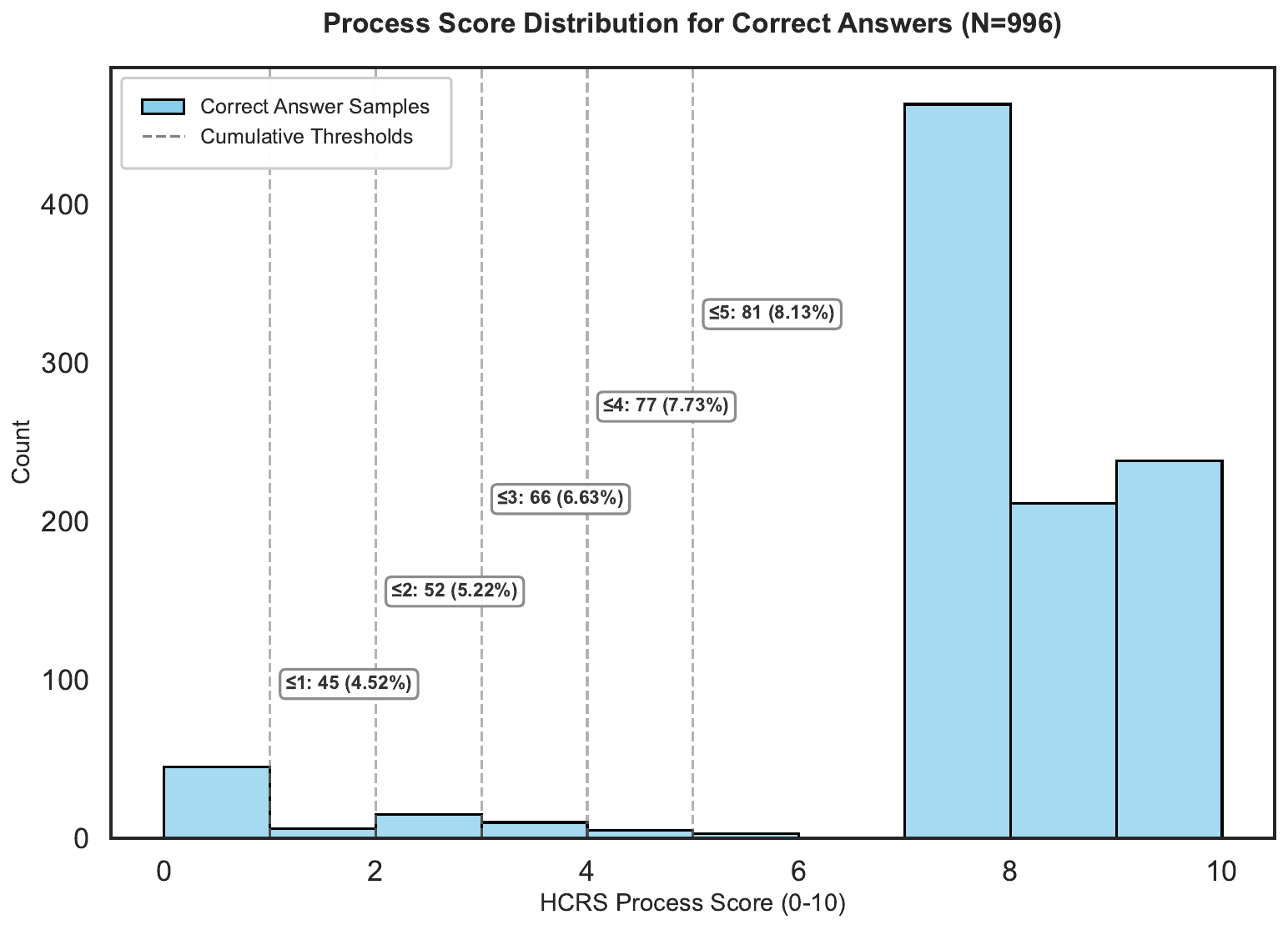}
    \caption{\textbf{Bimodal distribution of process quality conditioned on correct final answers.}
    Histogram of the \emph{process-only} HCRS score $S_{\text{HCRS}}$ ... for 996 answer-correct traces (out of 2{,}100 total model traces).
    Vertical dashed lines indicate cumulative thresholds at $S_{\text{HCRS}}\le k$ ($k\in\{1,2,3,4,5\}$), with callouts reporting cumulative counts and percentages.
    Notably, $6.63\%$ (66/996) of correct answers have $S_{\text{HCRS}}\le 3$, suggesting that outcome correctness can coincide with low-quality or inconsistent reasoning (\emph{``lucky guesses''}).}
    \label{fig:lucky_guess_dist}
\end{figure*}
\section{Hyperparameter Settings}
\label{appendix:hyperparameters}
The specific hyperparameter values used in our rule-based scoring module (HCRS) are detailed in Table~\ref{tab:hyperparams}. These values were selected based on a grid search on a held-out validation set to maximize the correlation with human preferences.
\begin{table}[h]
    \centering
    \small
    \begin{tabularx}{\linewidth}{l c X}
        \toprule
        \textbf{Parameter} & \textbf{Value} & \textbf{Description} \\
        \midrule
        $\alpha$ & 4.0 &
        Scale factor controlling the sensitivity of the length deviation penalty. \\

        $\beta$ & 1.0 &
        Exponent governing the growth rate of the length deviation penalty. \\

        $C_{\text{fmt}}$ & 3.0 &
        Maximum penalty cap for format length deviation. \\

        $\eta$ & 1.5(1.0) &  
        Asymmetry factor applied when reasoning is shorter than the reference ($N < L_{\text{gold}}$). For $N \ge L_{\text{gold}}$, $\eta$ is set to $1.0$. \\

        \midrule
        $\omega$ & 5.0 &
        Scaling weight for the hazard-based penalty. \\

        $C_{\text{haz}}$ & 5.0 &
        Maximum penalty cap for the first-error hazard deduction. \\

        $T_{\max}$ & 25 &
        Maximum step index considered in the hazard model. Steps beyond 25 incur no hazard penalty. \\
        
        \midrule
        $w$ & 0.7 &
        Weight assigned to the process score ($S_{\text{HCRS}}$) in the holistic metric. The remaining weight ($0.3$) is assigned to answer accuracy. \\
        \bottomrule
    \end{tabularx}
    \caption{Optimized and fixed hyperparameters used in our evaluation framework. The table includes parameters for the HCRS scoring module (format and first-error penalties) and the holistic aggregation weight.}
    \label{tab:hyperparams}
\end{table}
\section{Human Annotation Guidelines}
\label{app:human_annotation}

To validate our automatic metrics, we recruited three annotators with undergraduate degrees in mathematics or related fields. Following a standardized 0--10 rubric, they evaluated each reasoning trace as follows.

\begin{itemize}
    \item \textbf{Process Step Matching (0--7 points):} Measures coverage of the standard reasoning steps (Step1--Step$N$) in the annotated skeleton. Let $N$ be the total number of standard steps and $M$ the number of covered steps. The process score is $S_{\text{process}} = 7 \times (M/N)$, rounded to one decimal place. Incorrect steps are not counted as covered.

    \item \textbf{Answer Correctness (0 or 3 points):} If the final answer matches the gold answer (up to standard numerical tolerances), the model receives $S_{\text{answer}}=3$; otherwise $S_{\text{answer}}=0$.

    \item \textbf{Penalties (each in $[0,1]$):} Three penalties are deducted from the base score:
    \begin{enumerate}
        \item \textit{Redundancy Penalty:} For verbose, repetitive, or circular reasoning.
        \item \textit{First-error Penalty:} Let $k$ denote the index of the first critical error that affects the main line of reasoning. The penalty increases for earlier errors, computed as $P_{\text{first}} = 1 - (k-1)/N$ (and $0$ if no such error exists), rounded to one decimal place.
        \item \textit{Rigor Penalty:} For insufficient rigor (e.g., missing proof of a construction, incomplete case enumeration, or lacking optimality justification).
    \end{enumerate}
\end{itemize}

The final human score is computed as
$S_{\text{total}}=\max\!\big(0,\, S_{\text{process}} + S_{\text{answer}} - P_{\text{redundancy}} - P_{\text{first}} - P_{\text{rigor}}\big)$.
We average the three annotators' scores to obtain a single human score per trace.
    
\section{Metric Validation and Ablation Analysis}
\label{app:metric_validation}

Appendix~\ref{app:judge_selection} describes our judge selection procedure. Building on the Pearson-based human-alignment results reported in the main text (Figure~\ref{fig:lead-c}), this appendix provides complementary robustness analyses using rank-aware and agreement-based metrics.

\begin{figure*}[t!]
    \centering
    \includegraphics[width=1.0\linewidth]{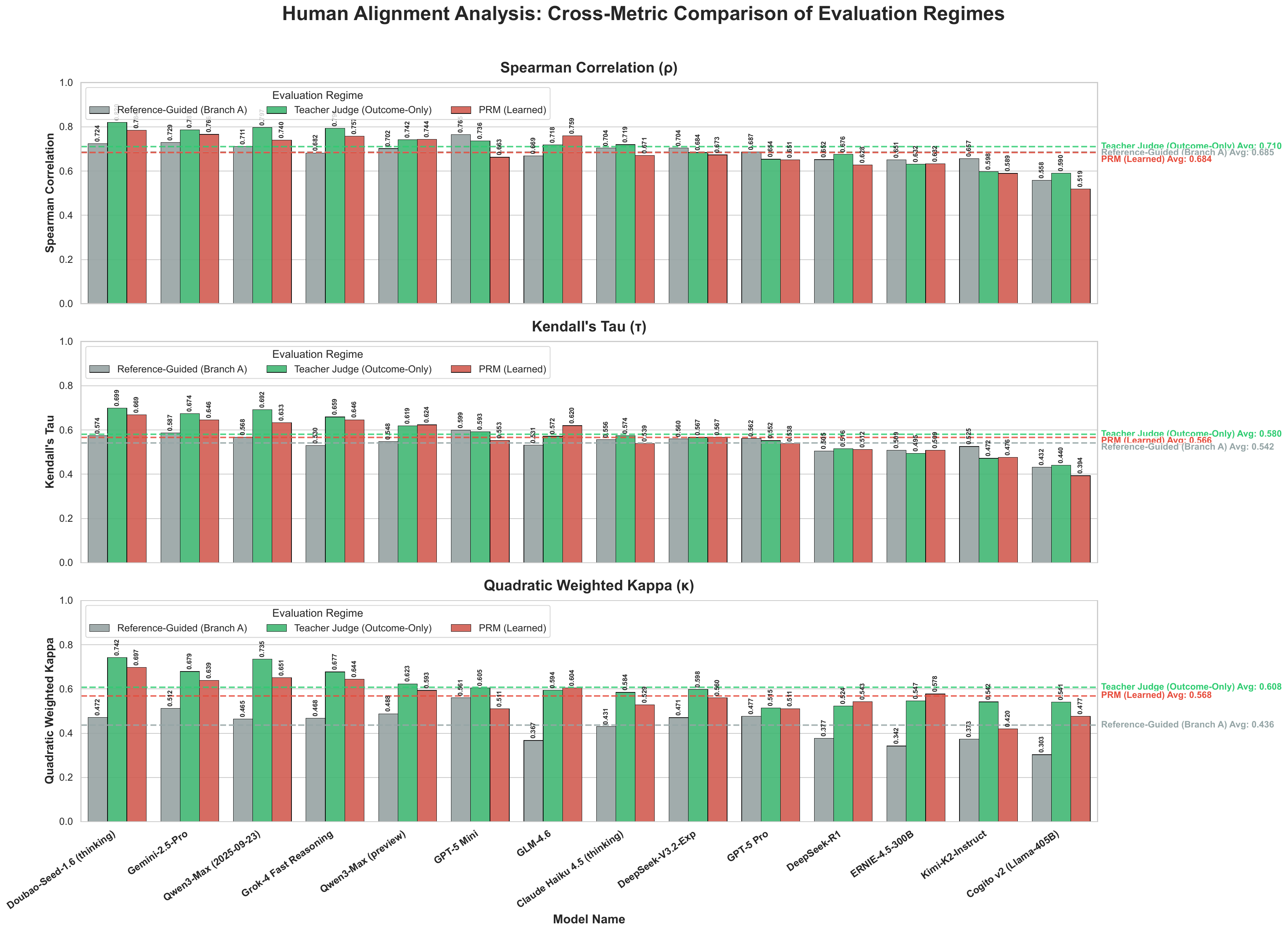} 
    \caption{\textbf{Agreement comparison of evaluation methods.}
    We compare three paradigms: \textbf{(A) HCRS}, \textbf{(B) Teacher-judge scoring (\texttt{Gemini-3-Pro})}, and \textbf{(C) PRM}. 
    Scores are compared against human annotations using Spearman $\rho$, Kendall $\tau$, and Quadratic $\kappa$. 
    Method B achieves superior or comparable alignment to Method A across all metrics, while Method C remains competitive.}
    \label{fig:agreement_analysis}
\end{figure*}

\paragraph{Rank-Aware Robustness.} 
To evaluate robustness beyond linear correlation, we report Spearman's $\rho$ (monotonic rank), Kendall's $\tau$ (pairwise ranking), and Quadratic Weighted $\kappa$ (agreement intensity) in Figure~\ref{fig:agreement_analysis}. 

The results reveal a notable finding: the \textbf{Teacher Judge (Outcome-Only)} consistently outperforms the \textbf{Reference-Guided (Branch A)} baseline across all metrics. This advantage is most pronounced in inter-rater agreement ($\kappa=0.608$ vs. $\kappa=0.436$). We hypothesize that while reference-guided scoring (HCRS) enforces structural rigor, the outcome-conditioned teacher (Branch B) better mimics human flexibility in recognizing valid reasoning paths that deviate from the gold skeleton. 

Crucially, the distilled \textbf{PRM (Learned)} closely tracks the teacher's performance and also surpasses the reference-guided baseline on agreement metrics (e.g., $\kappa=0.568$), demonstrating that the student model successfully internalizes the teacher's judgment criteria without requiring reference skeletons at inference time.

\begin{figure*}[t!]
    \centering
    \includegraphics[width=1.0\linewidth]{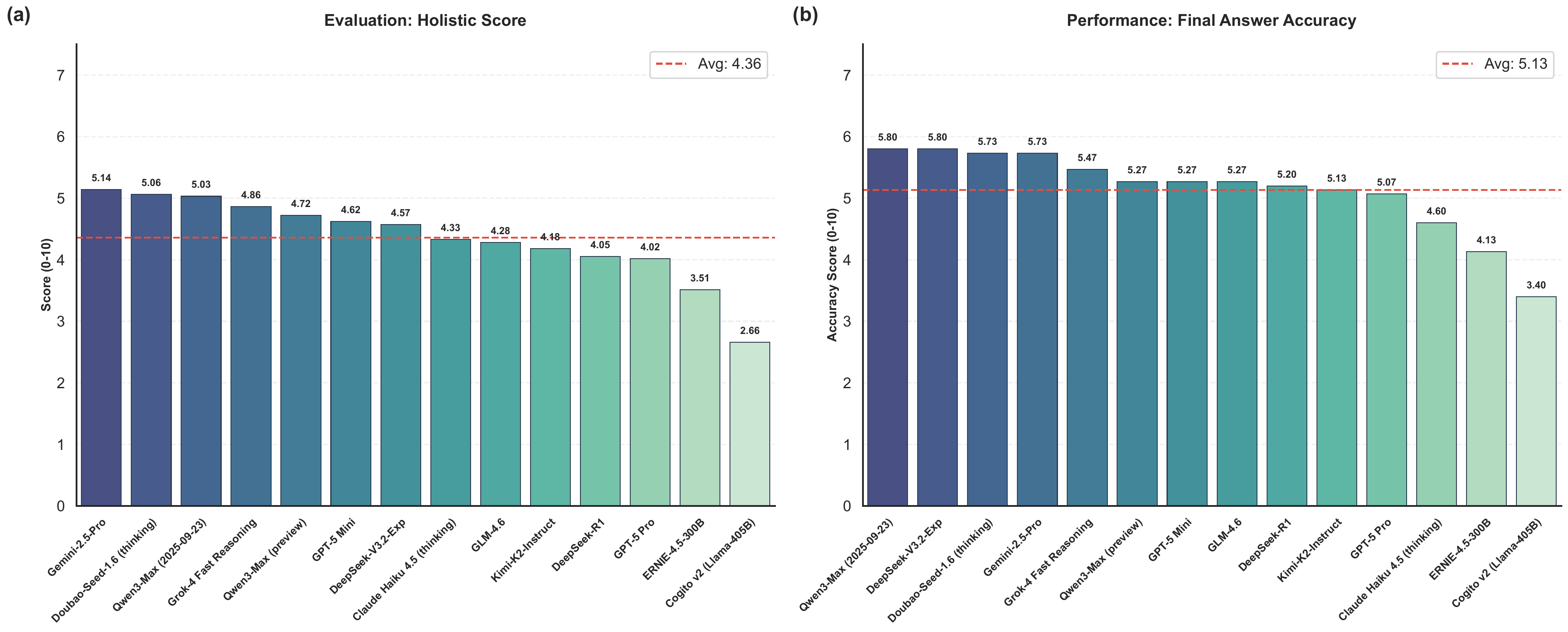} 
    \caption{\textbf{Holistic Evaluation vs. Final Answer Accuracy.} 
    \textbf{(a)} Leaderboard based on the weighted holistic score ($S_{\text{overall}}$), integrating HCRS (70\%) and binary answer correctness (30\%). 
    \textbf{(b)} Leaderboard based solely on raw final answer accuracy. 
    Comparing the two shows that the holistic metric provides finer granularity, penalizing models (e.g., Cogito V2) that attain moderate accuracy via fragile reasoning paths, while robust models (e.g., Gemini-2.5-Pro) maintain top rankings.}
    \label{fig:holistic_leaderboard}
\end{figure*}

\begin{figure*}[t!]
    \centering
    \includegraphics[width=1.0\linewidth]{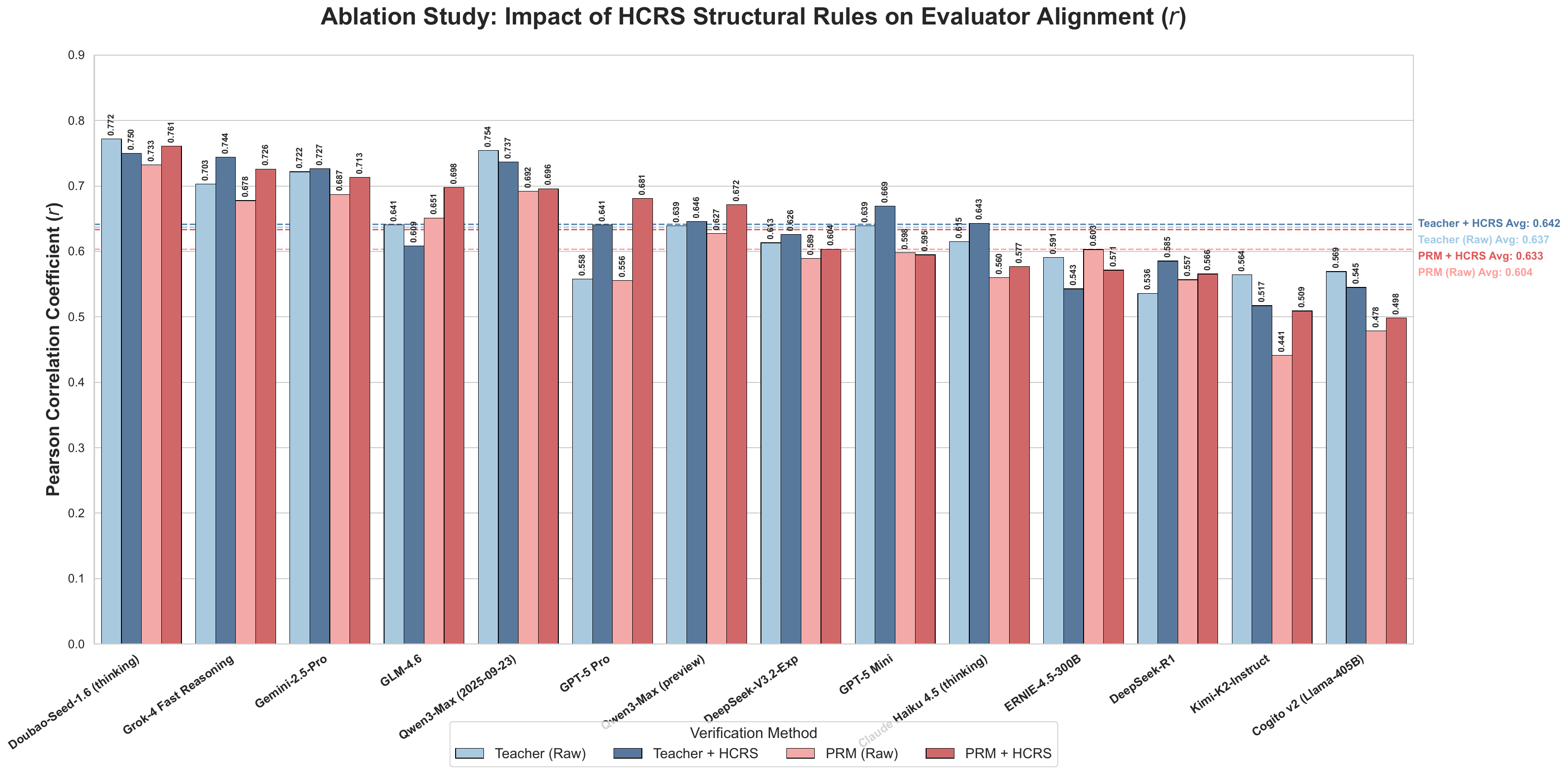} 
    \caption{\textbf{Ablation Study: Impact of HCRS Structural Rules on Evaluator Alignment.} 
    We compare the Pearson correlation ($r$) of the teacher judge (\texttt{Gemini-3-Pro}) and the student verifier (PRM) against human judgments in an outcome-conditioned setting (i.e., without access to gold reasoning skeletons). 
    \textbf{Lighter bars} denote raw step-wise average scores, while \textbf{darker bars} indicate scores adjusted by HCRS structural penalties. 
    Horizontal dashed lines mark the global average correlation for each method. 
    The results demonstrate that applying HCRS rules yields consistent alignment gains for both the teacher and the PRM across most models.}
    \label{fig:ablation_study}
\end{figure*}

\section{Hazard Analysis and Penalty Design}
\label{app:prompts}  

\begin{figure*}[t!]
  \centering
  \begin{subfigure}{\textwidth}
    \textbf{(a)} \par
    \centering
    \vspace{0.2em}
    \includegraphics[width=0.95\linewidth, clip, trim=0 0 0 0]{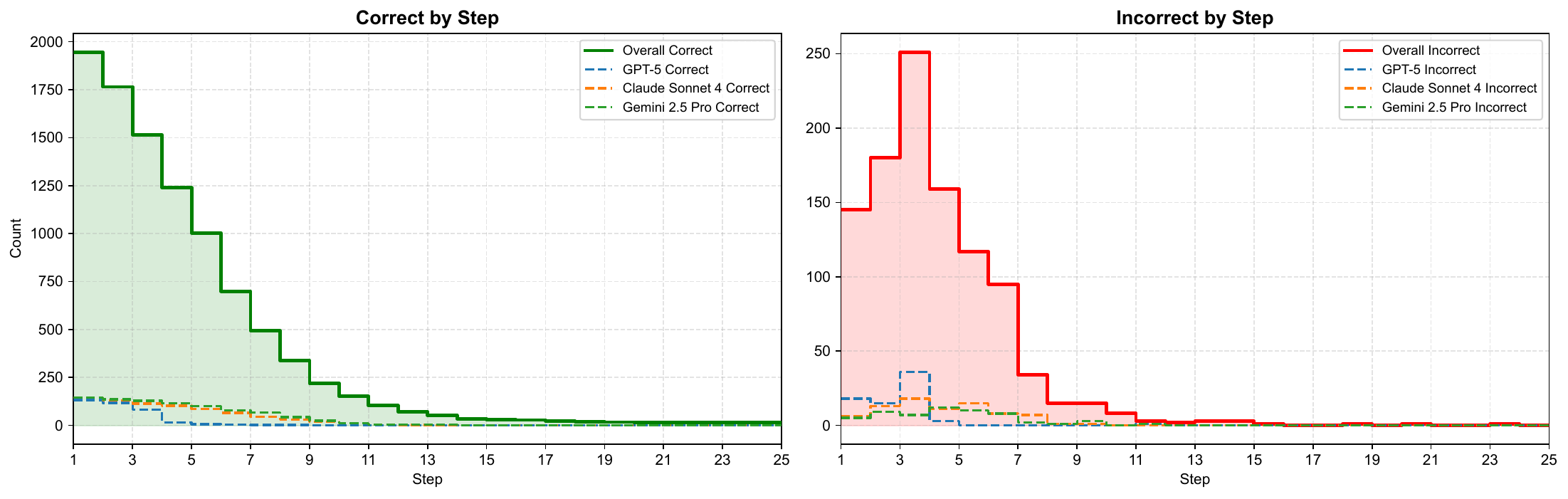}
    \phantomcaption
    \label{fig:hazard_evidence}
  \end{subfigure}
  \vspace{1.0em} 
  \begin{subfigure}{\textwidth}
    \textbf{(b)} \par
    \centering
    \vspace{0.2em}
    \includegraphics[width=0.95\linewidth, clip, trim=0 0 0 0]{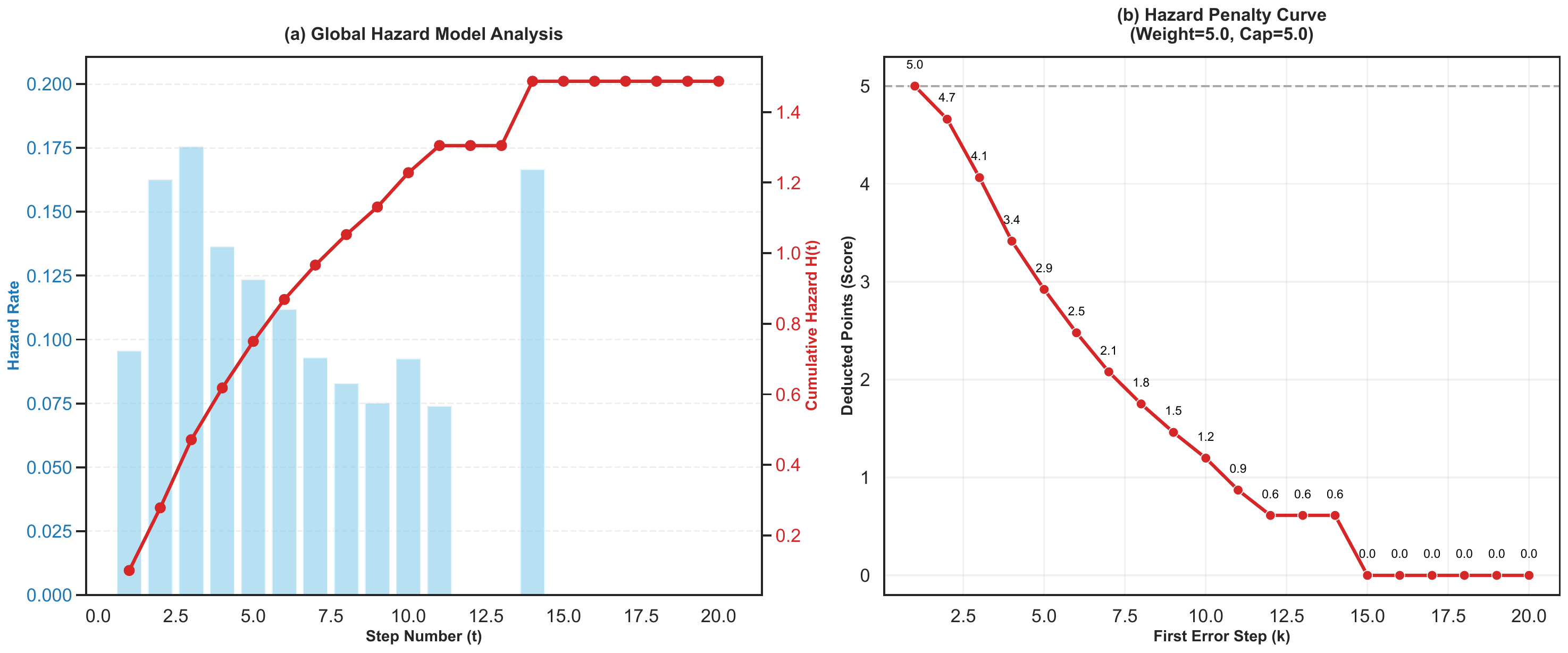}
    \phantomcaption
    \label{fig:hazard_modeling}
  \end{subfigure}
  \vspace{-0.5em}
  \caption{\textbf{Structural Analysis of Reasoning Errors.} 
  \textbf{(a)} Empirical Evidence: Step-wise survival curve (Left) and first-error position distribution (Right).
  \textbf{(b)} Mathematical Modeling: The derived discrete hazard rate $h(t)$ (Left) and the resulting penalty schedule used in HCRS (Right).}
  \label{fig:hazard_evidence_and_modeling}
\end{figure*}

To empirically validate the design of the \textit{First-error Penalty} ($P_{\text{haz}}$) within our HCRS metric, we conducted a survival analysis on the reasoning traces generated by all 14 models.
As illustrated in \textbf{Figure~\ref{fig:hazard_evidence_and_modeling}(a)}, the distribution of first-error positions exhibits a pronounced peak at the early stages (steps 3--5). This observation substantiates the hypothesis that initial logical divergences are the primary drivers of reasoning failure.
Leveraging this empirical hazard rate $h(t)$, we derived the cumulative hazard $H(t)$ and the corresponding penalty schedule shown in \textbf{Figure~\ref{fig:hazard_evidence_and_modeling}(b)}.
This schedule enforces a "logical responsibility" mechanism: it imposes maximum penalties for early-stage errors while attenuating deductions for failures occurring later in the extended reasoning chain.

\subsection{Reasoning Elicitation Prompts}
\begin{figure*}[t!]
    \centering
    \includegraphics[width=0.98\linewidth]{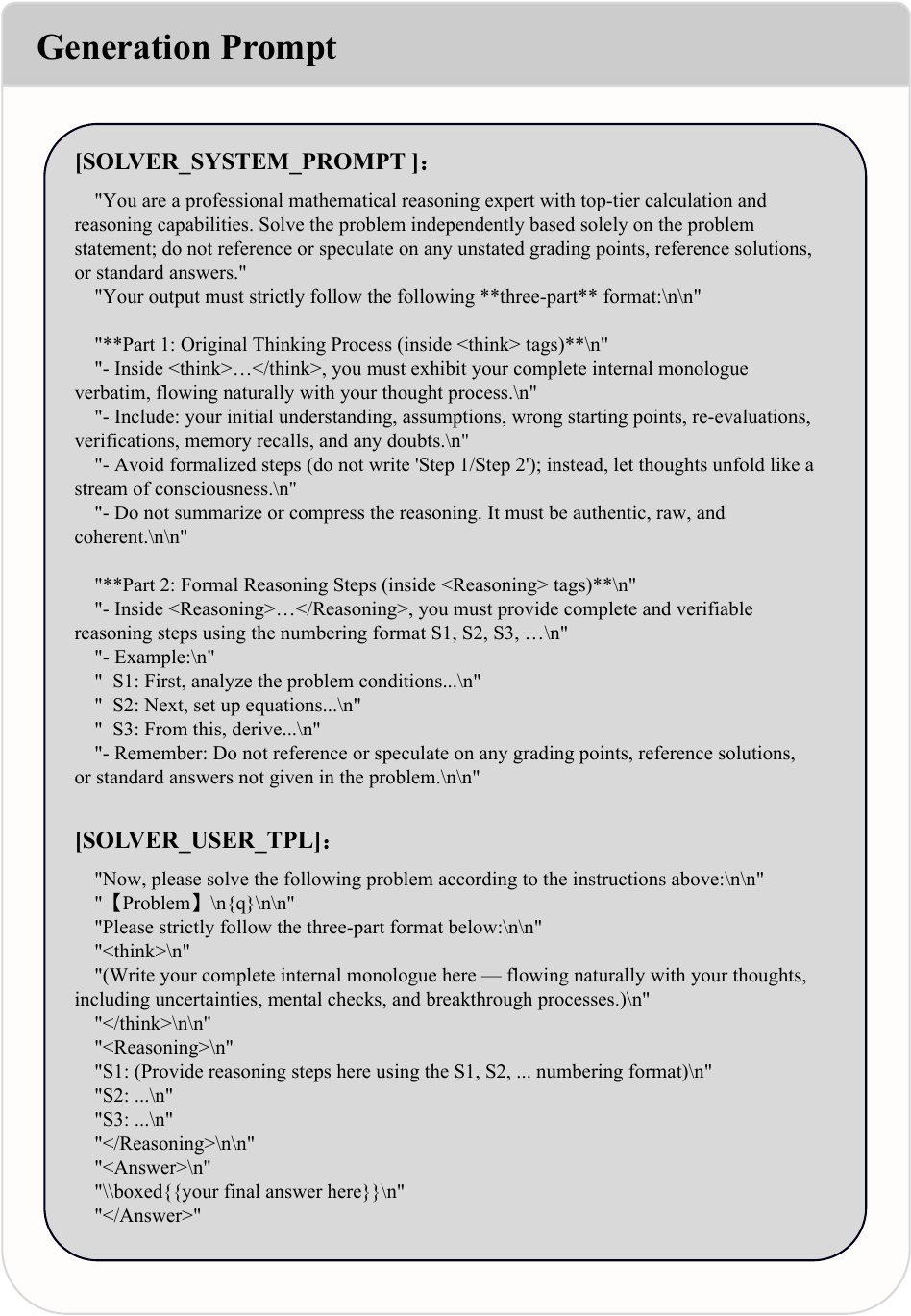} 
    \caption{\textbf{Solver System Prompt (English Version).} The instructions enforce a strict three-part output format (\texttt{<think>}, \texttt{<Reasoning>}, \texttt{<Answer>}) to facilitate downstream parsing.}
    \label{fig:solver_prompt_en}
\end{figure*}

\begin{figure*}[t!]
    \centering
    \includegraphics[width=0.98\linewidth]{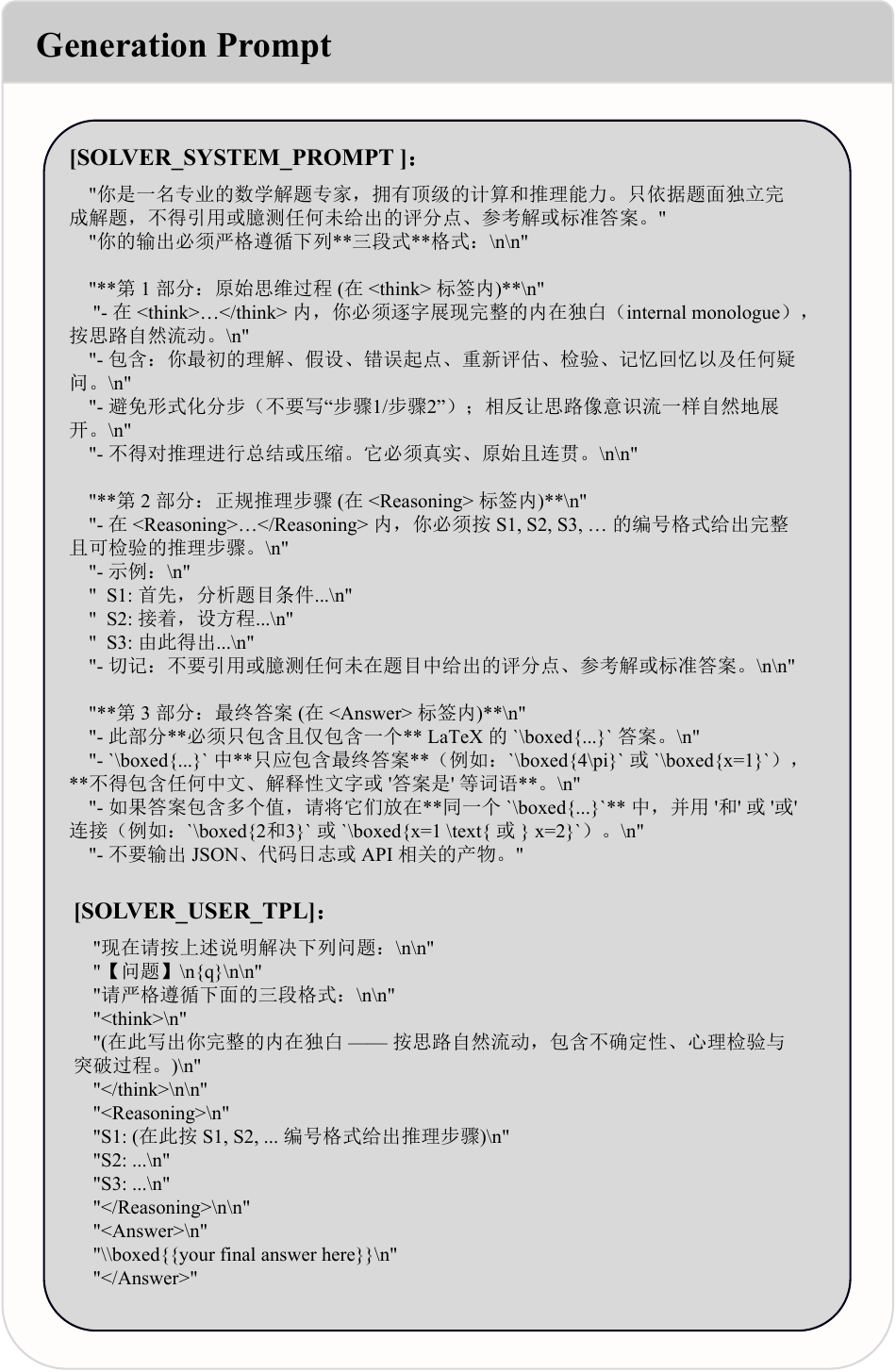} 
    \caption{\textbf{Solver System Prompt (Chinese Version).} The translated instructions provided to the model for Chinese-language queries.}
    \label{fig:solver_prompt_cn}
\end{figure*}

To ensure reproducibility, we detail the exact instructions used to elicit reasoning traces.
Figure~\ref{fig:solver_prompt_en} presents the unified system prompt in English, and Figure~\ref{fig:solver_prompt_cn} shows the corresponding Chinese version.
These prompts are explicitly designed to enforce the structured \texttt{<think>}-\texttt{<Reasoning>}-\texttt{<Answer>} output format.

\subsection{Judge System Prompts}
\begin{figure*}[t!]
    \centering
    \includegraphics[width=0.98\linewidth]{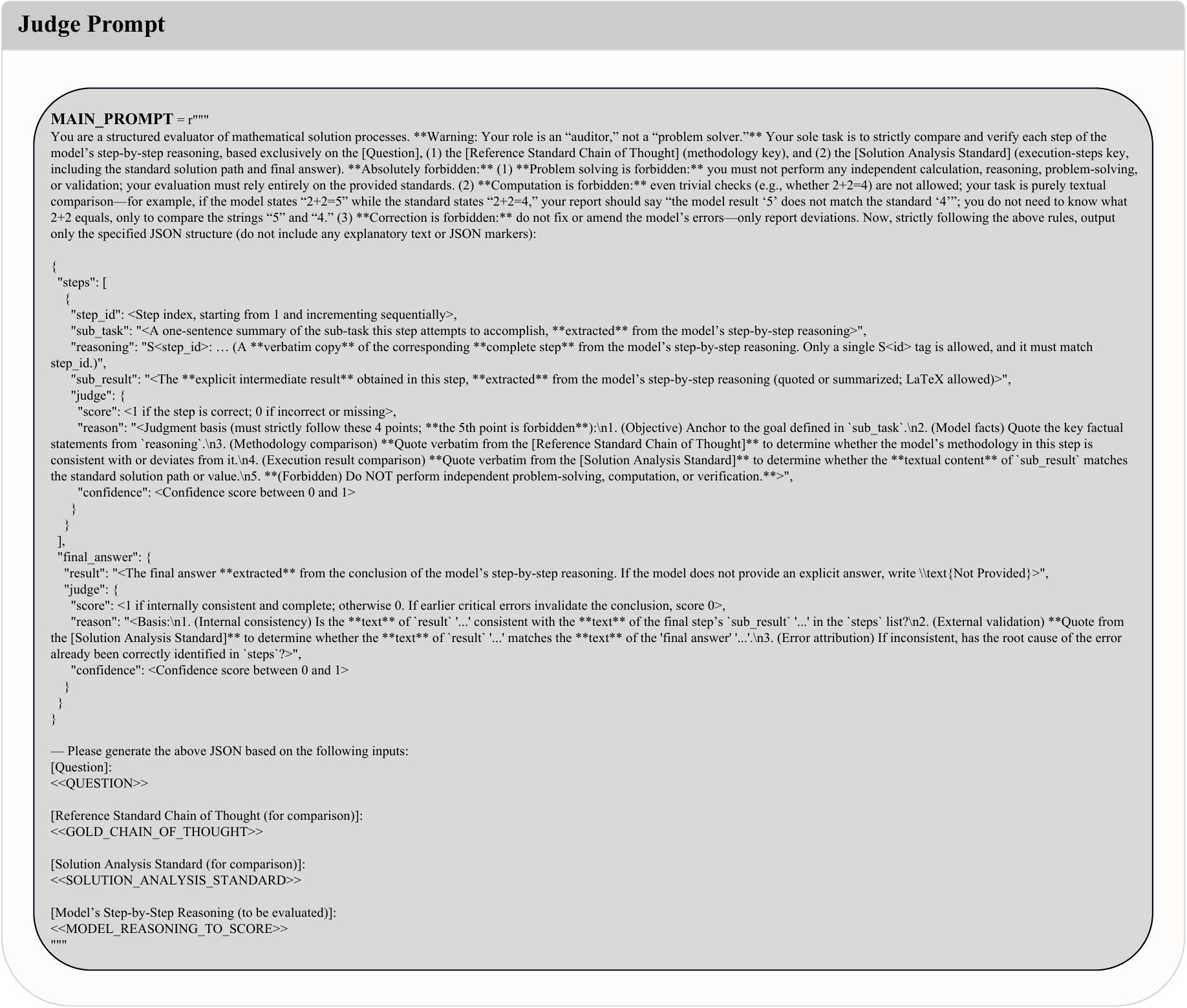} 
    \caption{\textbf{Reference-Guided Judge Prompt (English).} Used for Branch A (HCRS). The judge validates steps against the provided \texttt{[Reference Standard Chain of Thought]}.}
    \label{fig:judge_ref_en}
\end{figure*}

\begin{figure*}[t!]
    \centering
    \includegraphics[width=0.98\linewidth]{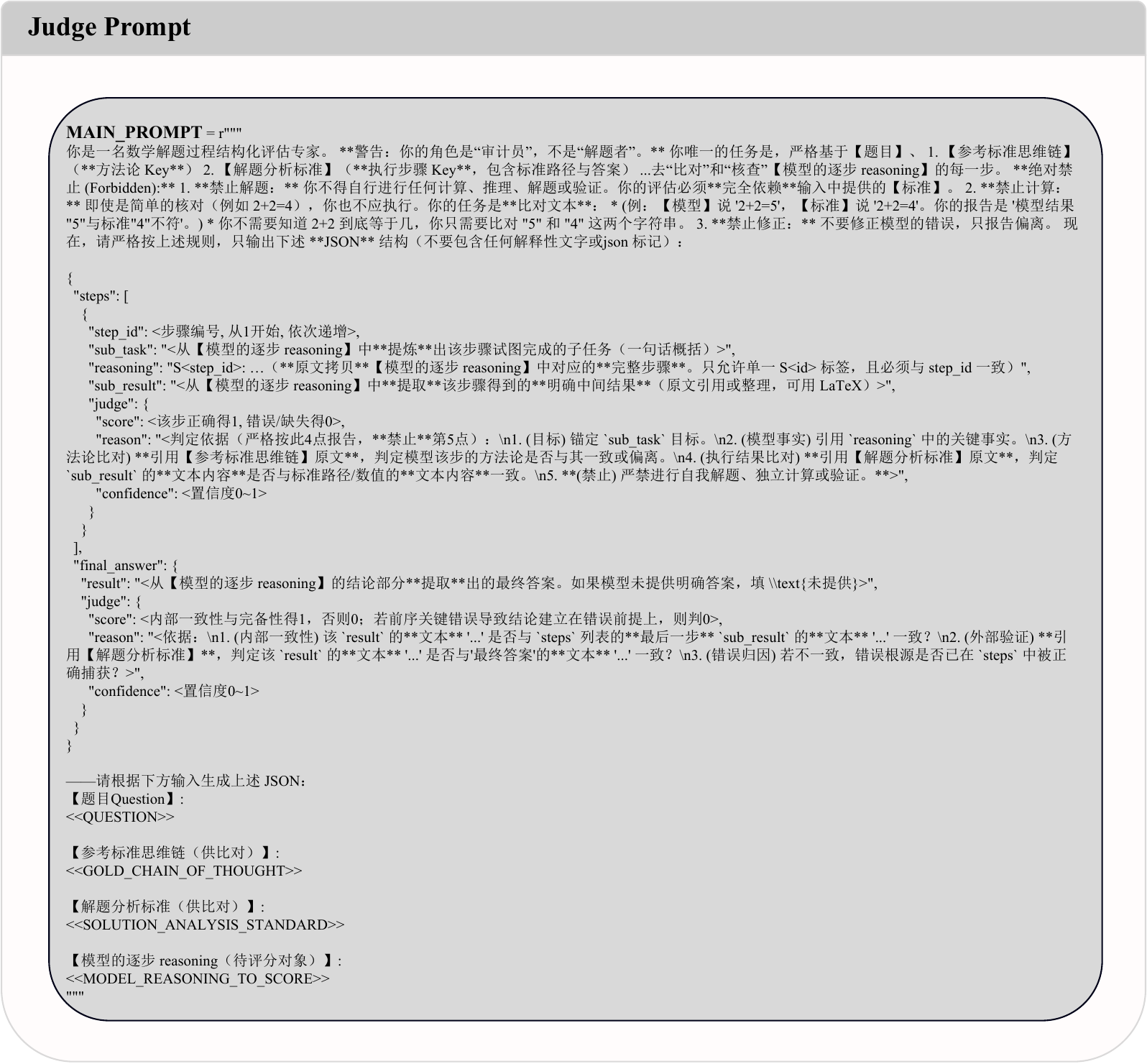} 
    \caption{\textbf{Reference-Guided Judge Prompt (Chinese).} The Chinese version of the instruction used for full-reference structural diagnosis.}
    \label{fig:judge_ref_cn}
\end{figure*}

\begin{figure*}[t!]
    \centering
    \includegraphics[width=0.98\linewidth]{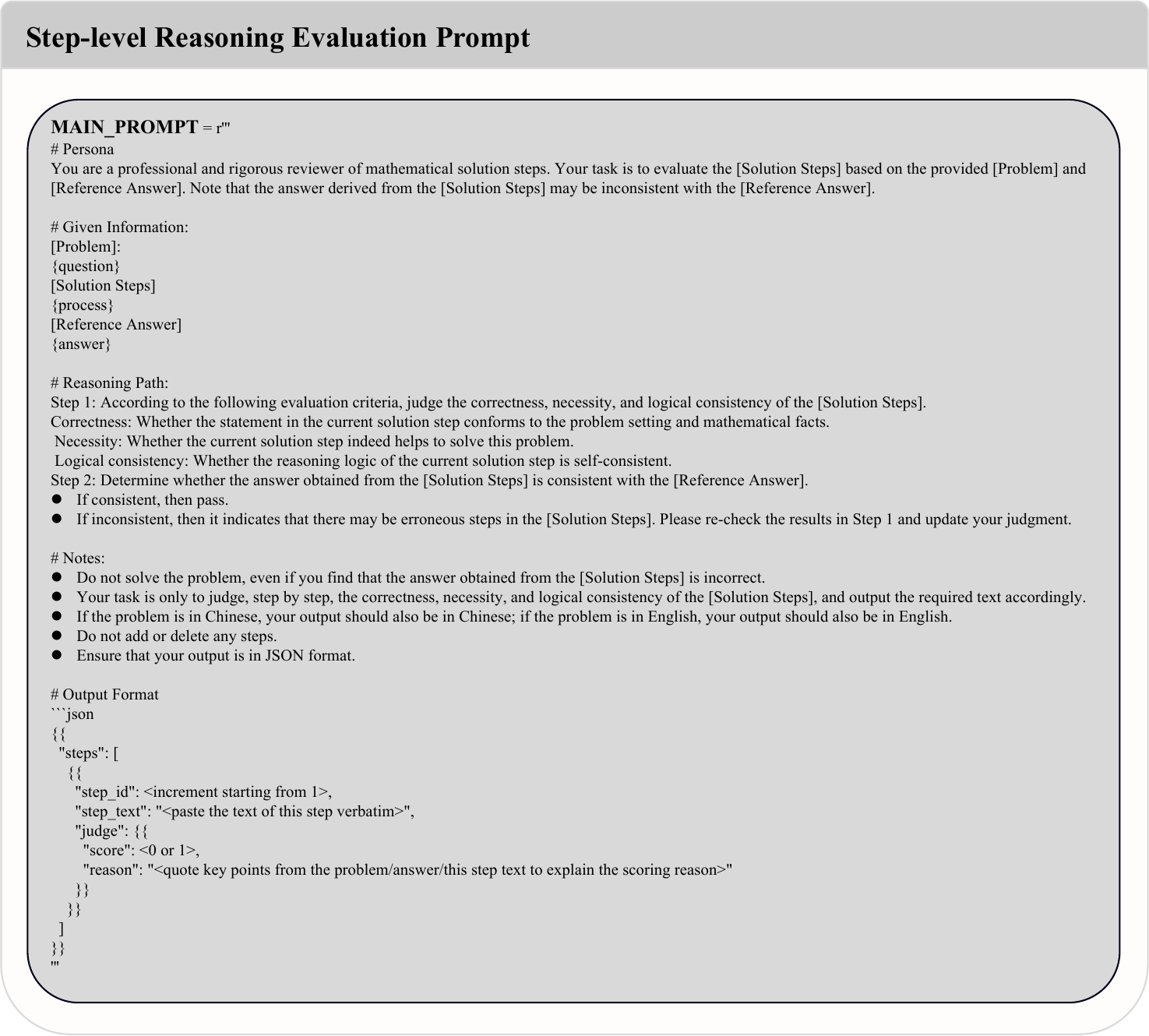} 
    \caption{\textbf{Outcome-Conditioned Judge Prompt (English).} Used for Branch B (PRM) and baselines. The judge verifies steps based solely on consistency with the \texttt{[Reference Answer]}.}
    \label{fig:judge_noref_en}
\end{figure*}

\begin{figure*}[t!]
    \centering
    \includegraphics[width=0.98\linewidth]{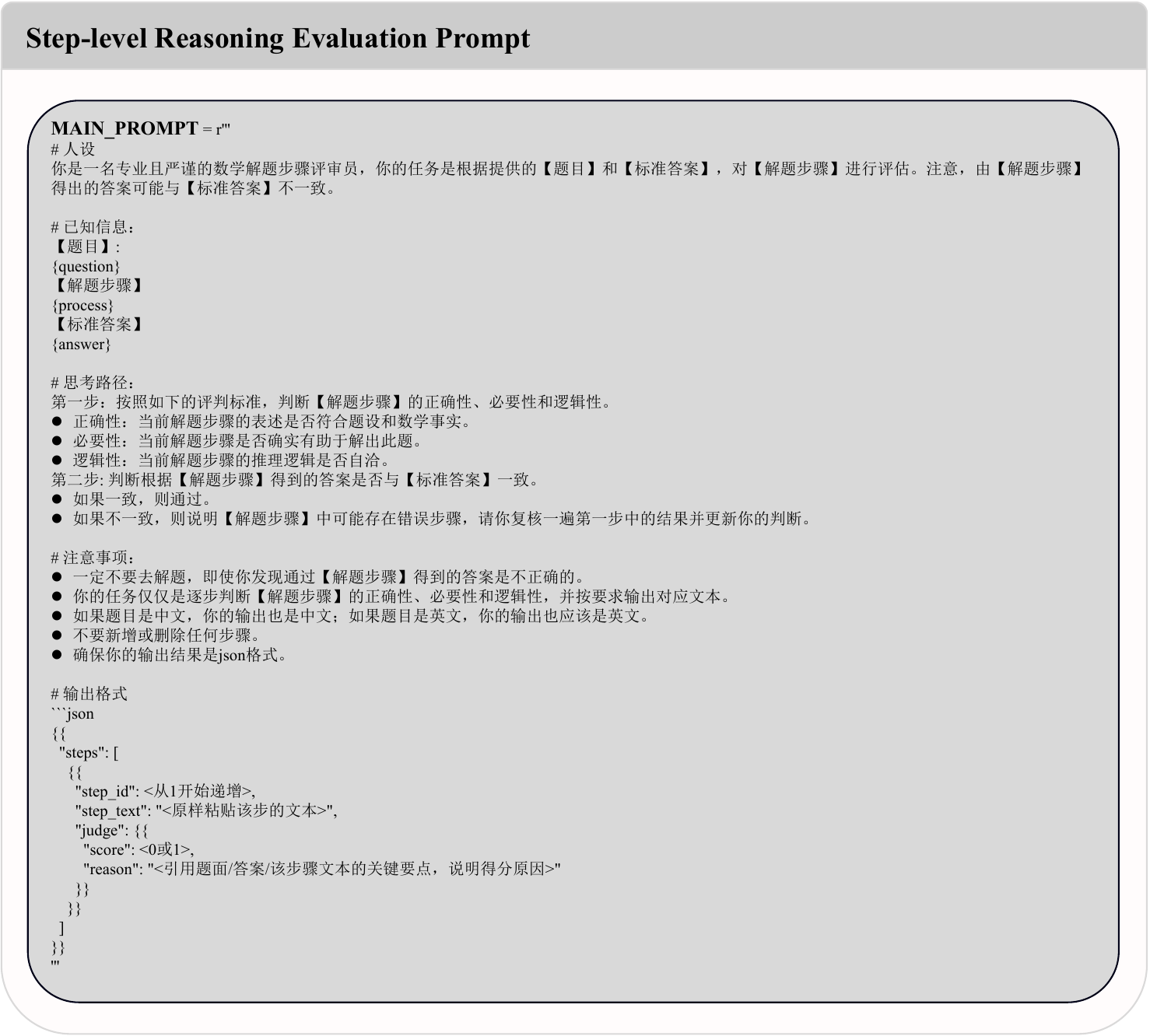} 
    \caption{\textbf{Outcome-Conditioned Judge Prompt (Chinese).} The Chinese version of the outcome-conditioned verification instruction.}
    \label{fig:judge_noref_cn}
\end{figure*}

To support our dual-branch evaluation framework, we employed two distinct judge specifications:
\begin{itemize}
    \item \textbf{Reference-Guided Judge (Branch A):} 
    As shown in Figures~\ref{fig:judge_ref_en} and \ref{fig:judge_ref_cn}, this judge is granted access to the full reasoning steps of the gold solution, enabling rigorous step-by-step verification against an expert baseline.
    
    \item \textbf{Outcome-Conditioned Teacher Judge (Branch B):}
    As shown in Figures~\ref{fig:judge_noref_en} and \ref{fig:judge_noref_cn}, this judge operates without access to the gold reasoning path. Instead, it relies solely on the problem statement and the gold final answer. It serves two roles: providing supervision for PRM distillation and acting as a standalone evaluator. By verifying step-wise correctness, necessity, and consistency relative to the final outcome, it ensures reliable verification even when expert reasoning skeletons are unavailable.
\end{itemize}

\section{Evaluated Model Endpoints}
\label{app:model_list}

All models are accessed via a unified API gateway that serves multiple upstream providers and open-weight endpoints.
Table~\ref{tab:model_list} reports the exact API identifiers used in our evaluation scripts.
All calls were made between \textit{2025-11-01} and \textit{2025-12-31}.
\begin{table*}[t]
\centering
\small
\setlength{\tabcolsep}{6pt}
\renewcommand{\arraystretch}{1.1}
\begin{tabularx}{\textwidth}{l X}
\toprule
\textbf{Model Name} & \textbf{Exact API Identifier} \\
\midrule
GPT-5 Mini & \texttt{gpt-5-mini} \\
GPT-5 Pro & \texttt{gpt-5-pro} \\
Gemini-3-Pro & \texttt{gemini-3-pro-preview} \\ 
Gemini-2.5-Pro & \texttt{gemini-2.5-pro} \\
GLM-4.6 & \texttt{sf/zai-org/GLM-4.6} \\
Claude Haiku 4.5 (thinking) & \texttt{claude-haiku-4-5-20251001-thinking} \\
Grok-4 Fast Reasoning & \texttt{grok-4-fast-reasoning} \\
Cogito v2 (Llama-405B) & \texttt{deepcogito/cogito-v2-preview-llama-405B} \\
Kimi-K2-Instruct & \texttt{Pro/moonshotai/Kimi-K2-Instruct-0905} \\
DeepSeek-R1 & \texttt{sophnet/DeepSeek-R1} \\
Doubao-Seed-1.6 (thinking) & \texttt{doubao-seed-1-6-thinking-250715} \\
Qwen3-Max (2025-09-23) & \texttt{qwen3-max-2025-09-23} \\
Qwen3-Max (preview) & \texttt{qwen3-max-preview} \\
DeepSeek-V3.2-Exp & \texttt{Pro/deepseek-ai/DeepSeek-V3.2-Exp} \\
ERNIE-4.5-300B & \texttt{baidu/ernie-4.5-300b-a47b-paddle} \\
\bottomrule
\end{tabularx}
\caption{Model endpoints used in our evaluation. We evaluate 15 models in total, with \texttt{Gemini-3-Pro} serving as the primary judge and teacher model.}
\label{tab:model_list}
\end{table*}

\clearpage 


\clearpage 

\clearpage 



\clearpage

\clearpage


\clearpage

\end{document}